\documentclass[12pt]{article}
\usepackage[margin =1.1in]{geometry}            
\usepackage{array}
\usepackage[hyphens]{url}
\usepackage{setspace,graphicx,bbm}
\usepackage{breakurl,booktabs} 
\usepackage[breaklinks]{hyperref}
\hypersetup{
     colorlinks   = true,
     allcolors    = blue
}
\newcommand\MyBox[2]{
  \fbox{\lower.9cm
    \vbox to 1.3cm{\vfil
      \hbox to 1.7cm{\hfil\parbox{1.7cm}{#1\\#2}\hfil}
      \vfil}%
  }%
}
\usepackage{multirow,rotating}
\usepackage[dvipsnames]{xcolor}
\usepackage[parfill]{parskip}    
\usepackage{amssymb,amsmath}
\usepackage[round]{natbib}
\usepackage{mathtools}
\usepackage{setspace,bbm}
\DeclareMathOperator*{\argmin}{argmin}

\title{\bf Modeling Dynamic User Interests:\\ A Neural Matrix Factorization Approach}

\author{
 Paramveer S. Dhillon\\
  University of Michigan, Ann Arbor\\
  \texttt{\color{blue}dhillonp@umich.edu}
 \and
  Sinan Aral\\
  Massachusetts Institute of Technology\\
  \texttt{\color{blue}sinan@mit.edu}
}

\date{}
\begin{document}
\maketitle
\begin{abstract}
{
In recent years, there has been significant interest in understanding users' online content consumption patterns. But, the unstructured, high-dimensional, and dynamic nature of such data makes extracting valuable insights challenging. Here we propose a model that combines the simplicity of matrix factorization with the flexibility of neural networks to efficiently extract nonlinear patterns from massive text data collections relevant to consumers' online consumption patterns. Our model decomposes a user's content consumption journey into nonlinear user and content factors that are used to model their dynamic interests.  This natural decomposition allows us to summarize each user's content consumption journey with a dynamic probabilistic weighting over a set of underlying content attributes. The model is fast to estimate, easy to interpret and can harness external data sources as an empirical prior. These advantages make our method well suited to the challenges posed by modern datasets. We use our model to understand the dynamic news consumption interests of Boston Globe readers over five years. Thorough qualitative studies, including a crowdsourced evaluation, highlight our model's ability to accurately identify nuanced and coherent consumption patterns. These results are supported by our model's superior and robust predictive performance over several competitive baseline methods.\\

{\bf Keywords:} {\it Machine Learning; Deep Learning; Natural Language Processing; Digital Marketing; User Profiling.}
}

\end{abstract}

\pagebreak
\section{Introduction}

The advent of the Internet and digitization of consumer activity has provided a golden opportunity for companies to gather more information about customers. Digital platforms can use the abundant clickstream data collected from consumers for a variety of purposes. For instance, they can track how consumers interact with their website and accordingly make adjustments to improve the user experience to maintain a sustained level of user engagement. They can also use consumer data to make product recommendations~\citep{bodapati2008recommendation}, assess the churn probability and customer lifetime value~\citep{moe2003buying,moe2004dynamic}, generate dynamic personalizations~\citep{hauser2009website,urban2013morphing}, offer customizations~\citep{ansari2003customization}, target prices~\citep{dube2017scalable}, target advertisements~\citep{goldfarb2011online,perlich2014machine}, and personalize search results~\citep{yoganarasimhan2016search}. Beyond just its business value~\citep{trusov2016crumbs,martens2016mining}, consumer data can also be leveraged for public policy ends.
The digital trails left by consumers on social media websites like Twitter can be used to gain insights into their psychological and physical well-being~\citep{schwartz2013characterizing,sinnenberg2017twitter}.

It should come as no surprise that consumer information is increasingly viewed as an essential strategic asset for companies. Despite or perhaps because of the exponential growth in data generation and collection over the past decade, generating actionable insights from this data faces three main challenges. First, online clickstreams and other user-generated content (UGC) often contains significant unstructured information which lives in very sparse and high-dimensional spaces.\footnote{By unstructured, we mean the kind of data that does not readily fit into a standard tabular format, e.g., text, image, audio, and video data. The usual way of encoding such data is via a one-hot-encoding. For example, in the case of text data such an encoding implies representing each word in English with a sparse vector of size the vocabulary of English ($\sim$ 300K) with all zeros, except a one at the location of the lexicographically sorted index of that word.} This makes statistical inference using traditional methods hard. Standard statistical inference methods typically estimate a parameter for each dimension and hence are unable to handle such an {explosion} of parameters efficiently. Second, the {dynamic} nature of this data further aggravates the challenge posed by sparsity owing to the inherent nonstationarity of the data generating process. However, it is this change in customer interests indicated by the dynamics of content consumption that is commercially very valuable to model since it may indicate purchase intent. Third, modeling user content consumption is an inherently different and more complex problem than the canonical problem of modeling purchase data commonly encountered in marketing. There is a finite assortment of products or items that customers can purchase from, e.g., clothes, books, soap, etc., however, when it comes to content consumption, users have access to an infinite assortment. For instance, there are no two online news articles that are the same. So, online content consumption is a domain where the assortment of products that customers can consume is always increasing, and there is little incentive to repeat a ``purchase.'' Hence, a key idea in modeling customers' content consumption is not to model the actual product, i.e., a specific news article, but instead, assume that each product is composed of a set of latent attributes and customers choose to consume those. For instance, those latent attributes can be news topics such as sports, politics, business, etc. Despite these challenges, companies have indeed managed to unlock some of the enormous potentials of textual data. Yet, it is clear that much remains untapped.

This paper proposes a novel neural matrix factorization framework for modeling dynamic user interests that addresses the above shortcomings. Our model refrains from directly modeling the actual content consumed (e.g., the specific news article or blog post) for the reasons just described but instead assumes that content is composed of a set of underlying latent attributes or factors. Each user's content consumption interests are then derived as a time-varying convex combination of these latent content factors. In a nutshell, our model factorizes a user's content consumption journey into a set of common content factors shared by all the users, and a set of user factors that define a user-specific dynamic weighting over the content factors. 

Since these user and content factors are estimated from the sparse and high-dimensional content that users consume, we develop a novel neural network architecture that allows us to efficiently extract nonlinear patterns from the content by learning flexible basis functions. Neural Networks have enjoyed immense success lately in learning flexible basis functions that adapt to the underlying data, thus enabling them to model complex nonlinear patterns in high-dimensional data such as text~\citep{goodfellow2016deep}. However, there is a concern regarding their ``black-box'' nature, which led us to combine neural networks with matrix factorization. The user and content factors estimated by our model lend interpretability to our results while still preserving the flexibility of neural networks.

Our approach is efficient to estimate and easily scales to large data sizes as it does not involve costly sampling procedures for model inference. It addresses the data sparsity issue by embedding the high-dimensional clickstream data into low-dimensional projections (also known as embeddings). As we will see later, these embeddings can be estimated in advance on an external data source; hence they act as an empirical prior and provide a source of statistical efficiency to our estimation approach. Our model handles dynamics efficiently by incorporating state dependence via a simple recurrent connection, which is temporally smoothed to provide robust regularized estimates of users' evolving interests. In summary, our model addresses the issues posed by sparsity and dynamics of large unstructured datasets and further models user interests over the latent content attributes as opposed to directly modeling the specific content item (news article) that was consumed.

We use our approach to model the dynamic news consumption interests of Boston Globe readers over several years. The latent factors estimated by our model are used to predict the content that users' will consume in the future as well as to generate interpretable trajectories of evolving user interests. The superior predictive performance of our model, coupled with the coherence of our latent factors as validated by crowdsourced user studies, highlights the potential of our approach as a news categorization, recommendation or user-profiling tool.

The rest of the paper is organized as follows. Next, we position our paper within the broader marketing and machine learning literature. Then, in Section 3, we provide an overview of the empirical setup of our problem and describe the data. We describe our model specifications in Section 4. Section 5 describes the results of our model estimation on content consumption data from Boston Globe. We discuss managerial implications and provide avenues for future research in Section 6.

\section{Related Work}

Our work contributes to several strands of literature. First, our work contributes to the marketing literature on modeling users' online consumption behavior. One of the earliest works in this area was by \citet{montgomery2004modeling}, who model the users' online behavior by analyzing their path on a major online bookseller's website. They build a dynamic multinomial probit model to predict purchase conversions. \citet{hui2009} considers a hybrid online-offline setting where they use data collected via RFID trackers to analyze in-store purchase conversions. This research on path analysis highlights some of the earliest efforts on using digital traces to predict managerially relevant decisions, but unlike this paper did not model the actual textual content consumed by the users. More recently, \citet{trusov2016crumbs} model the textual data consumed by users to generate user profiles by extending Correlated Topic Models (CTM)\textemdash a variant of Latent Dirichlet Allocation (LDA)~\citep{blei2003latent}. Their approach extends CTM to incorporate visitation intensity, heterogeneity, and dynamics and is tailored towards the task of behavioral ad-targeting. Methodologically, their approach relies on Markov Chain Monte Carlo (MCMC) sampling for model inference, which makes it slow to estimate, and the results highly sensitive to parameter initialization. Further, the complexity of their probabilistic model makes it difficult to incorporate even simple nonlinearities in the dependence between the users' interests and the text they consume. In contrast to that, our model is not only fast and efficient to estimate but can also easily incorporate flexible nonlinearities.

Next, our work contributes to the literature on modeling evolution of consumers' preferences and their sensitivities to various marketing variables. The most classic work in this area is by \citet{guadagni1983logit}, which models the evolution of brand preferences using exponential smooths of customer-level brand-loyalty parameters. Since then, there has been much follow-up work on modeling the evolution of brand preferences. More recently, \citet{dew2017dynamic} have used Gaussian processes to model the dynamics of consumer preferences. There has also been work on modeling nonlinear relationships between other marketing variables, e.g., advertising and sales~\citep{bruce2008pooling}, who used particle filters. Though this body of work is methodologically elegant and flexibly models consumer heterogeneity\textemdash a key construct in marketing\textemdash these approaches are computationally inefficient and rarely scale to large datasets. Further, these approaches are more tailored towards modelling physical products unlike our approach which models a digital product with an ever-increasing assortment\textemdash news articles.

Our work also contributes to the burgeoning literature in marketing using machine learning methods for studying customer interests using various forms of user-generated content (UGC). This literature uses multiple types of online feedback provided by users for instance in the form of consumer reviews, online chats or searches to model their interests~\citep{netzer2012mine,tirunillai2014mining,buschken2016sentence,liu2017consumers,timoshenko2019identifying}. Substantively, this work is closest to us in terms of modeling the latent structure in text. Our work is, however, different as it models the consumption of content as opposed to content generation by users via reviews, chats, or searches. In terms of methods, our work is significantly different from any of these approaches.  We propose a novel neural-network-based matrix factorization approach to model text data. The neural network component of our model allows us to incorporate flexible nonlinearities in our model. And, the matrix factorization formulation adds interpretability to our results akin to some of the probabilistic models mentioned above.

Finally, our work is also related to several matrix factorization-style models in machine learning, recommender system, and operations research literature. At a high-level, our model performs a similar matrix decomposition as done by Latent Semantic Analysis (LSA)~\citep{deerwester1990indexing}, by Latent Dirichlet Allocation~\citep{blei2003latent}\footnote{LDA is not a matrix factorization model. Still, it can loosely be considered a Bayesian version of LSA.} for document-term matrices, or by Hierarchical Poisson Factorization (HPF)~\citep{gopalan2015scalable} for implicit-feedback data. However, there are several critical differences, as we discuss in Section 4.4. Our work also extends some of the recent work on dynamic collaborative filtering~\citep{koren2009collaborative,xiong2010temporal} to settings in which the user feedback is not merely limited to clicks or ratings but also includes textual content. One of the recent works in the operations research literature by \citet{farias2016learning} also shares some methodological similarity with our work. It proposes a fast and efficient novel matrix factorization approach for learning user preferences from online activity trails. However, it is different in several critical aspects than our method. First, \citet{farias2016learning} is not interested in modeling the dynamics of user preferences, but instead, they model the traditional consumer funnel of search, browse, and purchase. Second, their approach is suited for products with a finite assortment such as online shopping, unlike news content in which the variety of products increases continuously. Finally, and most importantly, their approach doesn't model nonlinearities in consumption.


\section{Empirical Setting}
We model the dynamics of users' interests in the context of online news. Online news consumption is a perfect testbed for studying the evolution of user interests as a broad representative base of internet users consume content online. Further, news consumption patterns do often change saliently over time. For instance, there has been a substantial increase in interest in political news after the 2016 USA Presidential election. Similarly, there is an uptick in the consumption of news articles related to basketball or football during the playoff season. There are several reasons for these changes in news consumption patterns. They can change owing to customers' innate individual-level traits, for example, via self-discovery or learning about a new topic on the Internet. They can also fluctuate due to broad population-level trends, or they can change due to the variation in the availability of certain kinds of content in specific periods.

Studying these evolving dominant and niche characterizations of users' digital personas over a long time-period could provide insights into their equilibrium interactions with the news website. Modeling these news readership dynamics is also crucial from the perspective of content providers since it presents them with a plethora of personalization opportunities. Tapping into users' fluctuating tastes could allow content providers to optimize content placement on their website, for instance, via news categorizations tailored to a user's interests. It also opens up opportunities for personalized news stories through the news website itself or via a newsletter. Content personalization has been shown to increase reader engagement and customer lifetime value (CLV) and is, therefore, pivotal from a business standpoint.

Finally, the digital personas estimated from the dynamic readership patterns may be used for user profiling. User profiles concisely summarize a user's interests and have numerous digital marketing applications, including targeting advertisements. The model we propose in this paper uncovers such user profiles from raw user consumption data and can allow content providers to personalize content offerings.

\subsection{Data}
We use more than five-years worth of individual-level clickstream data from Boston Globe from February 1, 2014 to May 13, 2019 to perform our analysis. Globe\footnote{Website: \url{http://www.bostonglobe.com}} is one of the 25 largest newspapers by circulation in the USA. Our data contains fine-grained information about the users' online reading behavior and contains information such as which articles they read, how much time they spent reading those articles, and their subscription status. We further have access to granular demographic data for the visitors, such as area code, zip-code, device type (mobile or desktop), operating system, and country.

We perform our analysis at the week level since news stories typically last for a few days. Also, some people only read the news on weekends. So, one might not expect to see interesting dynamics in content consumption behavior on a day-to-day basis. Moreover, it is typical for users' interests to crystallize over time spans longer than a day. We further restrict our dataset by weeding out infrequent visitors\textemdash those who were active five times or less during our entire observation period. In other words, every user in our dataset visited the website at least five different times during our entire observation period from 2014-2019.

\begin{figure}
    \centering
    \includegraphics{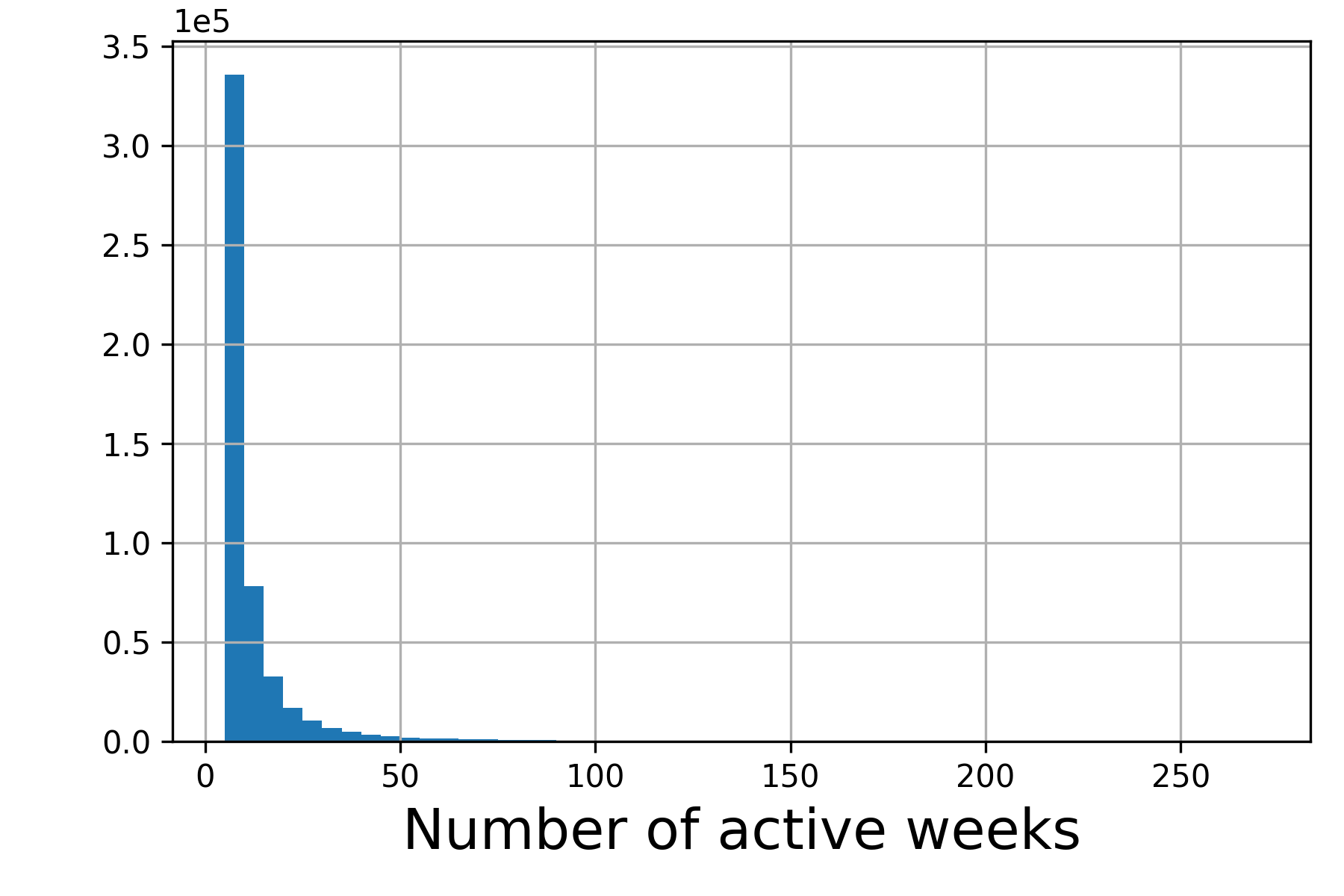}
    \caption{Frequency distribution of user activity.}
    \label{fig:user_activity}
\end{figure}

Our final dataset tracks 500,000 unique visitors over 276 weeks, leading to a total of 5,610,008 non-zero person-week observations.\footnote{Our full dataset contained a total of 11,399,021 unique users; however, due to computational/memory constraints, we randomly subsampled 500,000 users from the entire dataset. We were unable to estimate any bigger models with the computational resources at our disposal.} Of the total visitors, about 96.7\% were from the USA. Table~\ref{tab:data-summary} shows the summary statistics of our dataset. As can be seen, an average user made $1.64$ visits to the website each week and read $3.83$ articles. Further, an average user was active in $12.40$ weeks out of the entire 276 weeks, with a maximum of 264 and a minimum of 5. The frequency distribution of the number of weeks that the users were active is shown in Figure~\ref{fig:user_activity}.

Similar to other e-commerce businesses, Globe also counts each hit to its website as a unique visit, and a typical visit session lasts for 30 minutes. Hence, a visitor who spent 45 minutes on the site would have two visits attributed to them. Once a visitor clicks on a given news story, that article is counted as read. Globe's users fall into two categories: subscribers and anonymous visitors. Subscribers enjoy unfettered access to news and can be uniquely identified.  Anonymous visitors, on the other hand, are identified via cookies. If an anonymous visitor accesses the Globe website using two different browsers, then they would be counted as two unique users in our dataset. We understand that this is not an ideal scenario, but this is a shortcoming of all cookie-based digital fingerprinting schemes.

\begin{table}[htbp]
   \centering
     \begin{tabular}{l|cccc} 
      \hline
         & Min.& Median & Mean & Max.\\
      \hline
      \hline
      Visits per week   & 1&1&1.64& 626 \\
      News articles per week       &1&1&3.83&1400  \\
      Number of active weeks       &5&8&12.40&264  \\
      \hline
      \hline
   \end{tabular}
   \caption{Summary statistics of the visitation and reading behavior of the visitors to the Globe website. {\bf Note:} Our dataset consists of only those users who were active in at least 5 different weeks during our observation period.}
   \label{tab:data-summary}
\end{table}

\begin{figure}[h]
 \centering
    \includegraphics[width=.9\linewidth]{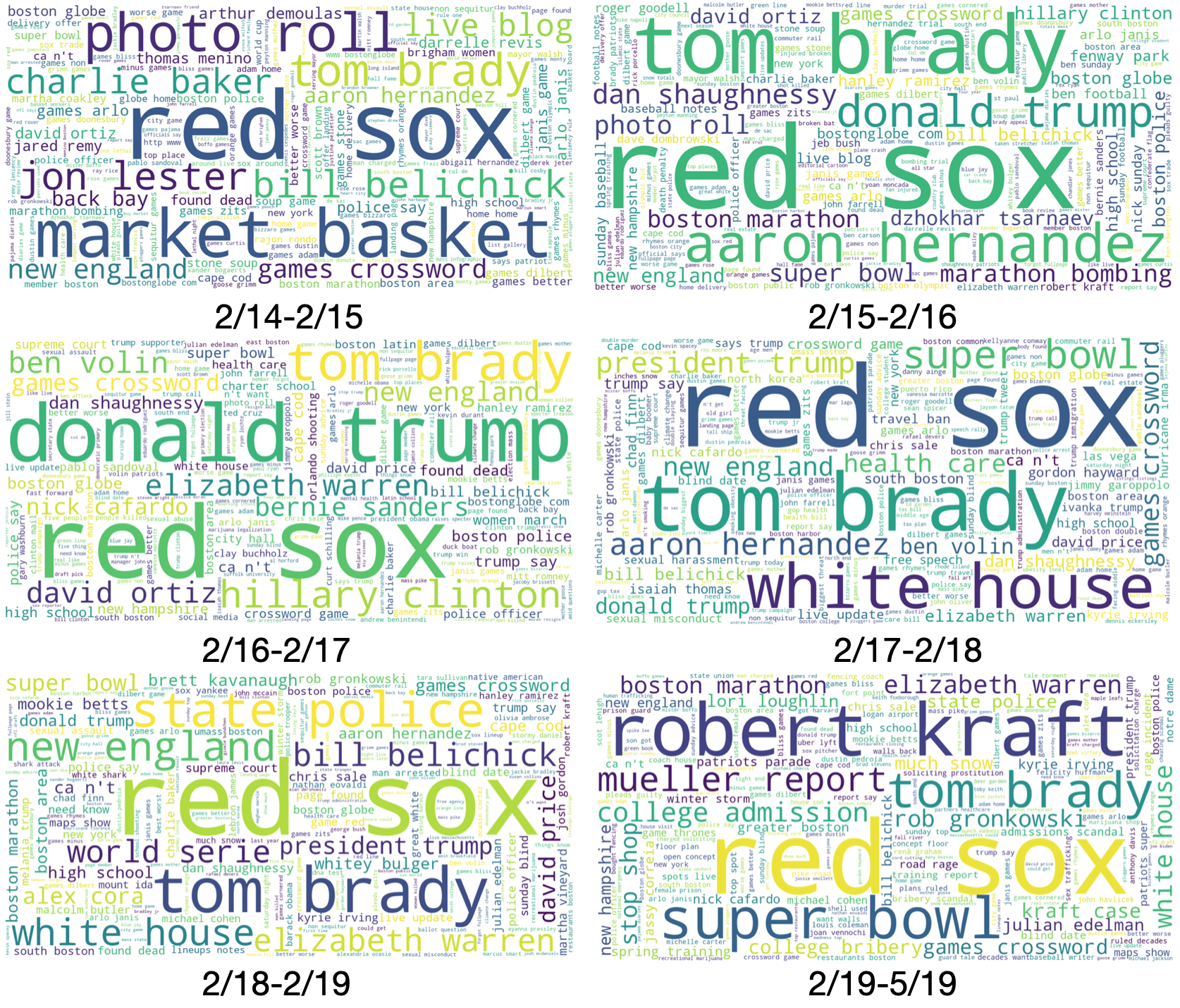}
\caption{Plot showing the prevalence of words in the news stories consumed by users in each 52 week (1 year) period starting February 2014. {\bf Note:} Larger font-size indicates the higher prevalence of those terms in users' consumption patterns.}
\label{fig:wordcloud}
\end{figure}

The textual component of our dataset consists of the headlines of the news stories that the users' read over the entire observation period. We do not use the actual body of the news story since users often choose to read an article just based on its headline. So, the headlines are predictive of users' content interests by themselves. Second, we excluded the body of the news stories due to computational issues as the headlines alone contained more than one hundred million words.
 We processed the news stories using the Natural Language Toolkit (NLTK)~\citep{bird2006nltk} by following a standard text-processing pipeline. We performed tokenization, lowercasing, and removal of stop-words. Our final processed text dataset consists of over one hundred million tokens (135,861,569) of text with a vocabulary (the number of unique words) of 85228. Figure~\ref{fig:wordcloud} plots the words in the news stories consumed by users broken down temporally. As expected, we can see the major sports and political events dominating consumption, but there is a high degree of heterogeneity in the nuanced consumption tastes of users. 
 
\section{Model}
Our model assumes that users have evolving latent content interests, and that they reveal a noisy version of these interests via the content they consume. So, we model the content consumed by users on Globe's website to infer their dynamic latent propensities for different types of content. It is accomplished in two steps. First, we assume that text content is composed of a set of underlying latent attributes that encapsulate general aspects of content that garner readers' interest. Next, each user's latent interests are modeled as a nonlinear time-varying weighting over these latent content attributes. Finally, we connect the users' interests across time to ensure a smooth evolution of their interests. These smoothed user-interest trajectories are then used to predict the users' future content consumption.

We propose a simple matrix factorization approach to decompose a user's content consumption traces into underlying latent content and user attributes. These user and content factors estimated by our model, in turn, lend interpretability to our model. Since these factors are learned from high-dimensional text data, we incorporate nonlinearities in these estimated factors via a novel neural network architecture to further boost their predictive ability. Before we delve into the details of our model, we introduce our notation and then provide an overview of matrix factorization more broadly for modeling content interests.

\subsection{Notation}
Let's denote the content consumed by user $i$ in time-period $t$ by  the column vector $x_i^t \in \mathbb{R}^{p\times 1}$. The column length $p$ represents the vocabulary size or the number of unique words in our dataset. In our case, $x_i^t$ denotes the set of words in the headlines read by the user $i$ at time $t$. The words are encoded using their one-hot encodings of size $p$, so if a word occurs more than once, the corresponding entry of the $x_i^t$ vector contains the count of that word. Further, let's assume that there are a total of $n$ users, and $\tau$ is the length of the observation period. Let's also assume that each user's unique identity is represented by an $n$ dimensional indicator vector $a_i$, i.e., a user-specific intercept. So, to summarize, our input data can be represented as $\tau$ slices of a $p$ dimensional column vector $x_i^t$ concatenated with a $n$ dimensional column vector $a_i$ to generate a $p+n$ dimensional column vector $z_i^t(=[x_i^t;a_i])$ for each user. Putting it all together and transposing the resulting matrix, our final input dataset consists of $\tau$ slices of $n\times (p+n)$ dimensional matrices $\{{ Z}\}^{t=1:\tau}$.

\subsection{Matrix Factorization for Modeling Users' Content Interests}
Our input data $\{{Z}\}^{t=1:\tau}$ can be seen as a type of interaction data where we observe the interactions of readers with content over time.\footnote{More precisely, the input data also contains $a_i$, which does not represent an interaction but describes user features. These user features can, though, also be assumed to be generated from the user factor.} So, a natural generative model for this data is to assume that each user $i$ is associated with a $K$ dimensional latent column vector $u_i^t$, and similarly, each word in the text $j$ is generated from a $K$ dimensional latent column vector $v_j$. This assumption is similar to the assumption about a word being generated from a $K$ dimensional topic made by Latent Dirichlet Allocation (LDA)~\citep{blei2003latent}. Next, we want to approximate the data matrix using the user and content factors that we assumed to have generated it as:
 \begin{equation}
   z_{ij}^t\approx v_j^{\top}u_i^t
   \label{eq:approx}
 \end{equation}
  where, ${\top}$ indicates matrix transpose. In the approximation given in Equation~\ref{eq:approx}, only the user factors $u_i^t$ change over time, whereas the content factors $v$ stay constant. Doing so permits a more parsimonious model; furthermore, there is no concrete reason to assume that the semantic representation of latent content factors drifts significantly over the observation period. Finally, the approximation described above can be recast as an optimization problem using a suitable loss function $\mathcal{L}(\cdot)$ as:
  
  \begin{equation}
      (U^t,V)=\argmin_{U^t,V}\mathcal{L}(Z^t,V^{\top}U^t)
       \label{eq:opt_prob}
  \end{equation}
  
  Recommender Systems literature has studied this optimization problem extensively~\citep{koren2009matrix,mnih2008probabilistic}. In that literature, the input data is usually the ratings, for instance, on a scale of one to ten, that the users give to items. Typically, these items have a finite and fixed assortment size, e.g., movies, or music, unlike news content that has an ever-growing assortment.  
  
  The optimization problem (Equation~\ref{eq:opt_prob}) requires a rank $K$ reconstruction of the data matrix ${Z}^t$ at each time step. We know that for squared-error loss function $\|Z^t-V^{\top}U^t\|_2^2$, the best such reconstruction is provided by the top $K$ eigenvectors of ${ Z}^t$~\citep{eckart1936approximation}. So, one can solve this optimization problem by computing the singular value decomposition (SVD) of the data matrix or it can be solved via iterative projection methods~\citep{seung2001algorithms,mairal2010online}. Alternatively, one can formulate an equivalent probabilistic version of this optimization problem by assuming Gaussian priors on the user and content factor matrices, and a conditional Gaussian distribution over the observed data as~\citep{mnih2008probabilistic} and compute its maximum a posteriori (MAP) estimate:
  \begin{eqnarray}
\nonumber
    p({Z}^t|{U}^t,{V},\sigma^2)&=&\prod_{i=1}^n\prod_{j=1}^{p+n}\mathcal{N}({z}^t_{ij}|{v}_j^{\top}{u}^t_{i},\sigma^2)\\
    \nonumber
    p({U}|\sigma_u^2)&=&\prod_{i=1}^n \mathcal{N}({u}_i^t|0,\sigma^2_u)\\
    \nonumber
    p({V}|\sigma_v^2)&=&\prod_{j=1}^{p+n} \mathcal{N}({v}_j|0,\sigma^2_v)
\end{eqnarray}
It turns out that maximizing the log-posterior of the above probabilistic model with the hyperparameters (i.e. the observation noise variance ($\sigma^2$) and prior variances ($\sigma_u^2, \sigma_v^2$)) kept fixed is equivalent to minimizing the sum-of-squared-errors objective function with
quadratic regularization terms shown in Equation~\ref{eq:sse}:
\begin{equation}
    (U^t,V)=\argmin_{U^t,V}\sum_{i=1}^n\sum_{j=1}^{p+n}\|z_{ij}^t-v_j^{\top}u_i^t\|_2^2 +\lambda_U \sum_{i=1}^n \|u_i^t\|_2^2 + \lambda_V \sum_{j=1}^{p+n} \|v_j\|_2^2
    \label{eq:sse}
\end{equation}
where $\lambda_U=\frac{\sigma^2}{\sigma_U^2}$ and $\lambda_V=\frac{\sigma^2}{\sigma_V^2}$. Much of the recommender systems literature has approached this problem in this fashion and optimized the biconvex objective function presented in Equation~\ref{eq:sse}, for instance, using Alternating Least Squares (ALS)~\citep{koren2009matrix}. This general matrix factorization framework is a bedrock of modern collaborative filtering approaches to recommendation in academia and industry. A variant of the above model also won the famous \$1 million Netflix Prize.\footnote{\url{https://en.wikipedia.org/wiki/Netflix_Prize}.}

In this paper, we extend this basic matrix factorization framework along three main dimensions to model the users' dynamic content consumption interests. 
 
\begin{enumerate}
    \item We incorporate nonlinearities into the user-specific latent factors. These nonlinearities are parameterized by a novel neural network architecture designed for the problem of modeling dynamics of content consumption. Neural Networks have enjoyed immense success in the recent past in extracting patterns from high-dimensional data by learning adaptive basis functions~\citep{goodfellow2016deep}. Hence, our neural network allows us to flexibly model the nonlinear dependence between the high-dimensional textual content and the users' latent interests.
    
    \item We introduce state-dependence between the latent user factors as ${u}_i^t$=f(${u}_i^{t-1}$). It is an important element of our model as prior research has shown strong evidence of habit formation in news consumption. We model this evolution of user tastes also via our neural network architecture. We connect the current and past estimates of the latent states of the user interests and then smooth them via exponential smoothing.
    \item We adapt the general matrix factorization (MF) framework presented in Equation~\ref{eq:sse} to the task of text modeling. MF has been used extensively in generating recommendations via collaborative filtering based on rating data. News articles are, however, inherently different than the ``items'' typically considered in the recommender systems literature as their assortment increases rapidly over time. That said, our approach does have connections to some of the text-modeling frameworks which loosely fit into the MF framework. We discuss those connections in detail in Section 4.4.
\end{enumerate}

In light of these, Equation~\ref{eq:sse} changes as:  
\begin{eqnarray}
\nonumber
\left[\{U\}^{t=1:\tau},{V},{\Theta}\right]&=& \argmin_{\{U\}^{t=1:\tau},{V},{\Theta}} \sum_{i=1}^n\sum_{j=1}^{p+n}\sum_{t=1}^{\tau}\|z^t_{ij}-\text{g}(v_j^{\top}u_i^t;{ \Theta})\|_2^2 + \lambda_U \sum_{i=1}^n\sum_{t=1}^{\tau} \|u_i^t\|_2^2\\ 
&&+ \lambda_V \sum_{j=1}^{p+n} \|v_j\|_2^2\;\;\text{such that}\;\; u^{t}_i=\text{f}(u_i^{t-1})\;\forall\;t=1:\tau
\label{eq:mat_fac_mod}
\end{eqnarray}
where $g(\cdot)$ encodes the neural network parameterized by ${\Theta}$ and $f(\cdot)$ represents the functional form of the state-dependence between the user interests.  Next, we describe our model in detail.

\subsection{Neural Network Architecture}
Our model is described by Equation~\ref{eq:mat_fac_mod}. We operationalize the nonlinearities and the state-dependence via a novel neural network architecture. Neural Nets have enjoyed remarkable success in the last decade in terms of providing state-of-the-art performances in several tough problems involving high-dimensional datasets such as those arising in speech, text, images, and video~\citep{murphy2012machine,goodfellow2016deep}. Further, Neural Nets allow us to flexibly incorporate families of nonlinearities, which is harder to accomplish with splines, kernel methods, or other nonlinear modeling techniques. A comprehensive introduction to Neural Nets is beyond the scope of this paper. We instead refer the reader to a popular textbook on this subject by \citet{goodfellow2016deep}.

The key behind the success of Neural Nets is their ability to learn superior data representations, and central to the notion of representation learning is the concept of an embedding~\citep{bengio2003neural,bengio2013representation}. An embedding is essentially a dense low-dimensional representational summary of a high-dimensional input such as text or an image. In our case, the embedding $e_w$ of a word $w$ is a map  $e_w: \mathbb{R}^{p} \to \mathbb{R}^{d}$, where $p$, as defined earlier, is the high-dimensional one-hot representation of a word and $d$ is the embedding dimensionality with $p\gg d$. Recall that $p$ is the number of unique words in our data. The one-hot encoding for a word, then, is just a vector of size $p$ with all zeros and just a one at the lexicographically sorted index of that word. The embedding dimensionality is the only new notation that we need to operationalize our model. The various modeling steps are described below.

\subsubsection{Embedding the Input Data.}
As the first step, we embed the high-dimensional input data \{$x_i^t$, $a_i$\} into $d$-dimensional spaces separately. This low-dimensional projection is performed via matrices ${E}_x$ and ${E}_a$, respectively. The embedding matrix ${E}_a$ is a model parameter and is estimated from the data. On the other hand, the matrix ${E}_x$, which embeds the words in the news headline, is {\it fixed}, and hence not estimated from the data. The embedding dimensionality $d$ is a hyperparameter of our model.

Embeddings capture generic properties of the high-dimensional input that they are projecting down to a low-dimensional space. So, the word embedding matrix ${E}_x$ encodes semantic information about the words that they are projecting down. Words with similar meanings are, therefore, closer in the embedding space~\citep{mikolov2013distributed,dhillon2015eigenwords}.
Hence, we can estimate ${E}_x$ on an independent dataset that is much larger than the size of our dataset, for instance, the entire Wikipedia, or all the English newswire. There are several such word embeddings trained on more massive datasets than ours that are publicly available, e.g., word2vec~\citep{mikolov2013distributed,mikolov2013efficient}, GloVe~\citep{pennington2014glove}, and Eigenwords~\citep{dhillon2011multi,tscca,dhillon2015eigenwords} among others. Using these ``pretrained'' word embeddings serves the purpose of a valuable empirical prior. However, no such pretrained embeddings are available for $a_i$ since the identity of users in our dataset is unique to our dataset and is not a general property that can be transferred from other datasets. Hence,  the embeddings ${E}_a$ need to be estimated from the data. The low-dimensional embedding of a given user $i$ can be simply obtained as ${E}_aa_i$.

In terms of the operationalization of pretrained word embeddings, we obtain the ``pretrained'' ${E}_x$ matrix as just described and {\it fix} it. Hence, these word embeddings are not estimated along with the rest of the model. The embedding of a specific word $w$ can then be retrieved as ${E}_xw$. The content consumed by users, $x_i^t$, however, consists of more than a single word, e.g., the headline {``GE unveils striking new headquarters for Fort Point.''} We obtain the embeddings for the entire sequence by retrieving the embeddings for individual words in the headline as ${E}_xx_i^t$ and then averaging them.

\subsubsection{Estimating a nonlinear hidden state for each user.}
A user's time-varying consumption and their unique identifier $a_i$ contribute to their latent state $\ell_i^t$ that represents their content interests at a given time step. So, once we have projected the inputs \{$x_i^t$,$a_i$\} to a $d$-dimensional space, we combine them nonlinearly to get the hidden state of that user at a given time step as shown in Equation~\ref{eq:hidden_state},

\begin{equation}
    \ell_i^t=\sigma_1({W}_{\ell}\cdot [ {E}_xx_i^t; {E}_aa_i])
    \label{eq:hidden_state}
\end{equation}
where ``;'' indicates row-wise concatenation. The nonlinear activation function is denoted by $\sigma_1(\cdot)$. We choose a Rectified Linear Unit (ReLU) as the nonlinearity for the sake of its simplicity and due to its lower susceptibility to the vanishing gradient problem~\citep{glorot2011deep}. A ReLU activation function is operationalized as: $\sigma_1(x)=\max(0,x)$. The ReLU nonlinearity is parameterized by the matrix ${W}_{\ell}$, which is a model parameter that is estimated from the data. The hidden state $\ell_i^t$ obtained after the nonlinear transformation is a $d$-dimensional column vector.

\subsubsection{Incorporating dynamics by combining a user's current and previous hidden states.}
Since the users' interests evolve, their final hidden state $u_i^t$ depends not only on the current inputs but also on the hidden state from the previous time step. We allow $u_i^t$ to depend nonlinearly on $\ell_i^t$ and $u_i^{t-1}$ as:

\begin{equation}
    u_i^t=\sigma_2({W}_u\ell_i^t+{W}_ru_i^{t-1})
    \label{eq_user_factor}
\end{equation}
The nonlinear transformation is parameterized by the matrices ${W}_u$ and ${W}_r$, both of which are estimated from the data. The output dimensionality of the user factor $u_i^t$ is $K$, where $K$ can be thought of as latent content attributes.  The role of $K$ in our model is analogous to the number of topics in a topic model such as Latent Dirichlet Allocation (LDA)~\citep{blei2003latent}. It is a model hyperparameter, and we show the robustness of our results to different choices of $K$.

Since the user factor $u_i^t$ captures $K$ different tastes of the user at that time step, it is natural that they represent probabilities and hence sum-to-one. Hence, we employ the softmax function as the nonlinearity $\sigma_2(\cdot)$ here. Softmax normalizes real-valued numbers into probabilities over the $K$ different content interests. It is operationalized as $\sigma_2({\bf z})_i=\frac{\exp(z_i)}{\sum_{j=1}^K\exp(z_j)}$, for $i=1,\ldots, K$ and ${\bf z}=(z_1,\ldots, z_k)$.

One would expect that a user's interests evolve gradually and smoothly. For instance, it is uncharacteristic for a user to be consuming content with {\it emotional valence} up to a certain time and then never engaging it with again. So, one issue with our operationalization shown in Equation~\ref{eq_user_factor} is that it doesn't ensure that user interest trajectories are smooth, and it turns out empirically that indeed they are choppy. We borrow an idea from the time-series modeling and brand choice modeling~\citep{guadagni1983logit} literature to address this problem. We use exponential smooths of the hidden state vectors to obtain user factors that evolve smoothly. The degree of smoothing is controlled by the hyperparameter $\alpha$. In light of this modification, Equation~\ref{eq_user_factor} changes as:

\begin{equation}
    u_i^t=\alpha\cdot \left[ \sigma_2({W}_u\ell_i^t+{W}_ru_i^{t-1})\right] + (1-\alpha)\cdot u_{i}^{t-1}
    \label{eq:exp_smooths}
\end{equation}

\subsubsection{Combining the user and content factors.}
The user factor $u_i^t$ provides a probability distribution over a user's interest in the $K$ latent content attributes at time $t$.\footnote{$u_i^t$ may no longer be a probability distribution after performing exponential smoothing. So, we scale it to make it sum-to-one after the smoothing step.} The temporal snapshots of the user factor at different times give us the dynamics of their interests. The content factor, denoted by the matrix $V$, represents the words that constitute each of the $K$ content attributes. $V$ projects each of the $K$ latent content attributes to a $d$-dimensional space, the same low-dimensional space as the word embeddings. Hence, one can find the words that constitute each of the $K$ latent content attributes by finding the nearest neighbors of each row of the $V$ matrix from the word embedding matrix ${E}_x$. Finally, the user and content factors are combined to provide a noisy rank-$K$ and $d$-dimensional reconstruction of the original input $x_i^t$ as:

\begin{equation}
    r_i^t=V^{\top}u_i^t
\end{equation}

The reconstruction vector $r_i^t$ can be seen as a projection of the $K$-dimensional user factor $u_i^t$ onto the $d$-dimensional embedding space.

\subsubsection{Minimizing the Loss function.}
The content and user latent factors condense the content consumption of all the users into (1) $K$ content attributes shared by all the users encoded into the matrix $V$, and (2) a user's dynamic weighting over those $K$ content attributes which is embedded into the vector $u_i^t$. The vector $r_i^t$ obtained by multiplying $V$ and $u_i^t$ provides a reconstruction of the input data as it lies in the same d-dimensional space.

We define our loss function to minimize the discrepancy between the input $x_i^t$ and its reconstruction $r_i^t$. Our loss function can be seen as similar to the one used by Principal Component Analysis (PCA)~\citep{murphy2012machine} or autoencoders~\citep{goodfellow2016deep} as these methods also minimize reconstruction error. Trivially, our model can be seen as an encoder-decoder architecture also, where $u_i^t$ encodes the inputs into a fixed-length vector, and the decoder then decodes it into some destination format, e.g., a translated sentence in a new language~\citep{bahdanau2014neural}. However, the crucial difference is that in our case, the source and destination are the same as we're reconstructing the input itself at each time step. Hence, our model is a recurrent autoencoder.

We optimize a squared-error loss function for the sake of simplicity and due to some recent results showing its superior performance on various text, image, and speech tasks~\citep{hui2020evaluation}. The optimization problem is shown in Equation~\ref{eq:loss_min}. A key observation that can be made is that fixing the word embeddings ${E}_x$ and making them non-trainable is important for making our model work.

\begin{equation}
   [\{U\}^{t=1:\tau},V]= \argmin_{{U}^{t=1:\tau},V} \sum_{i=1}^n\sum_{t=1}^{\tau} \|{E}_xx_i^t-r_i^t\|_2^2
   \label{eq:loss_min}
\end{equation}

\begin{figure}[htbp]
   \centering
   \includegraphics[width=0.9\textwidth]{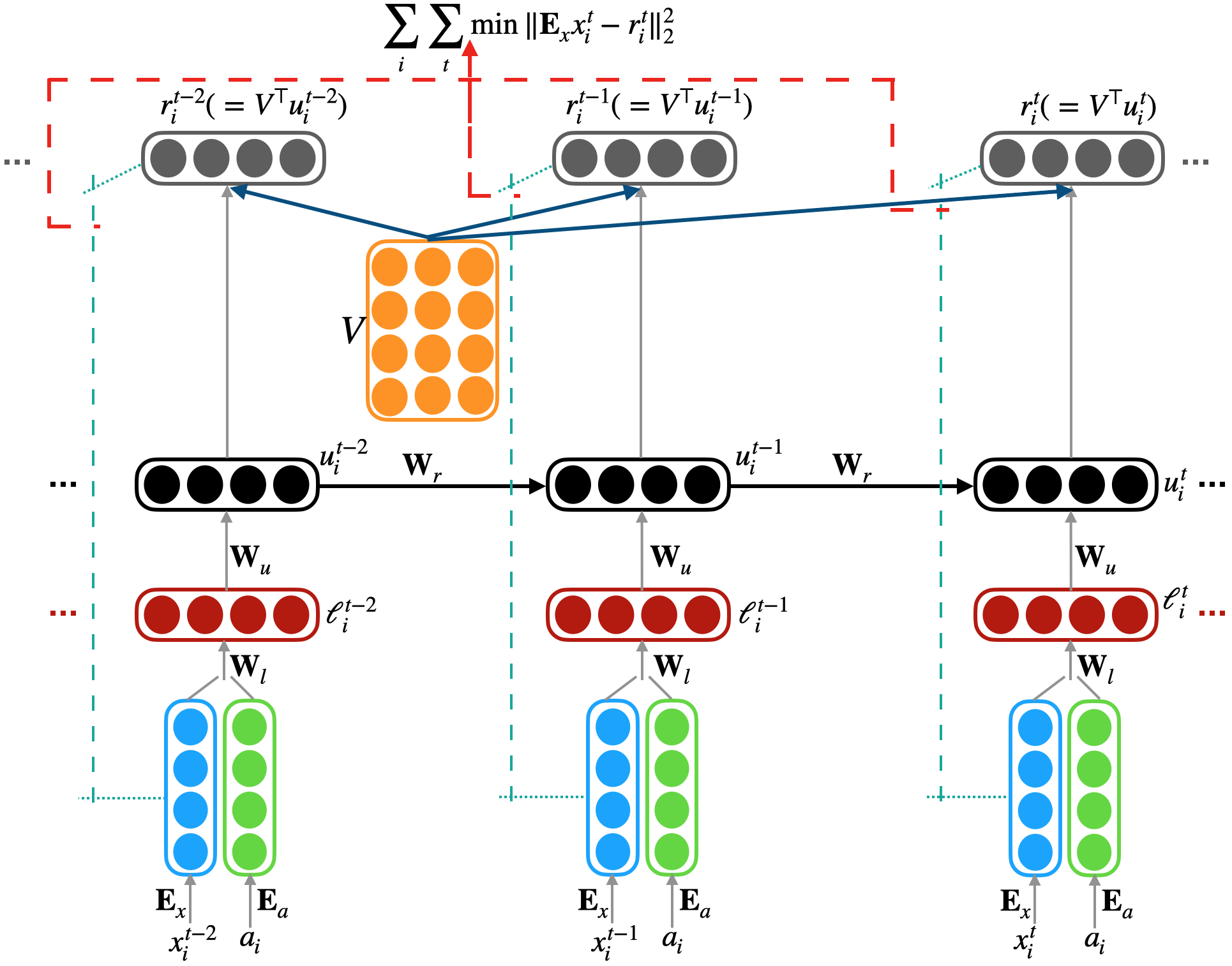} 
   \includegraphics[width=0.5\textwidth]{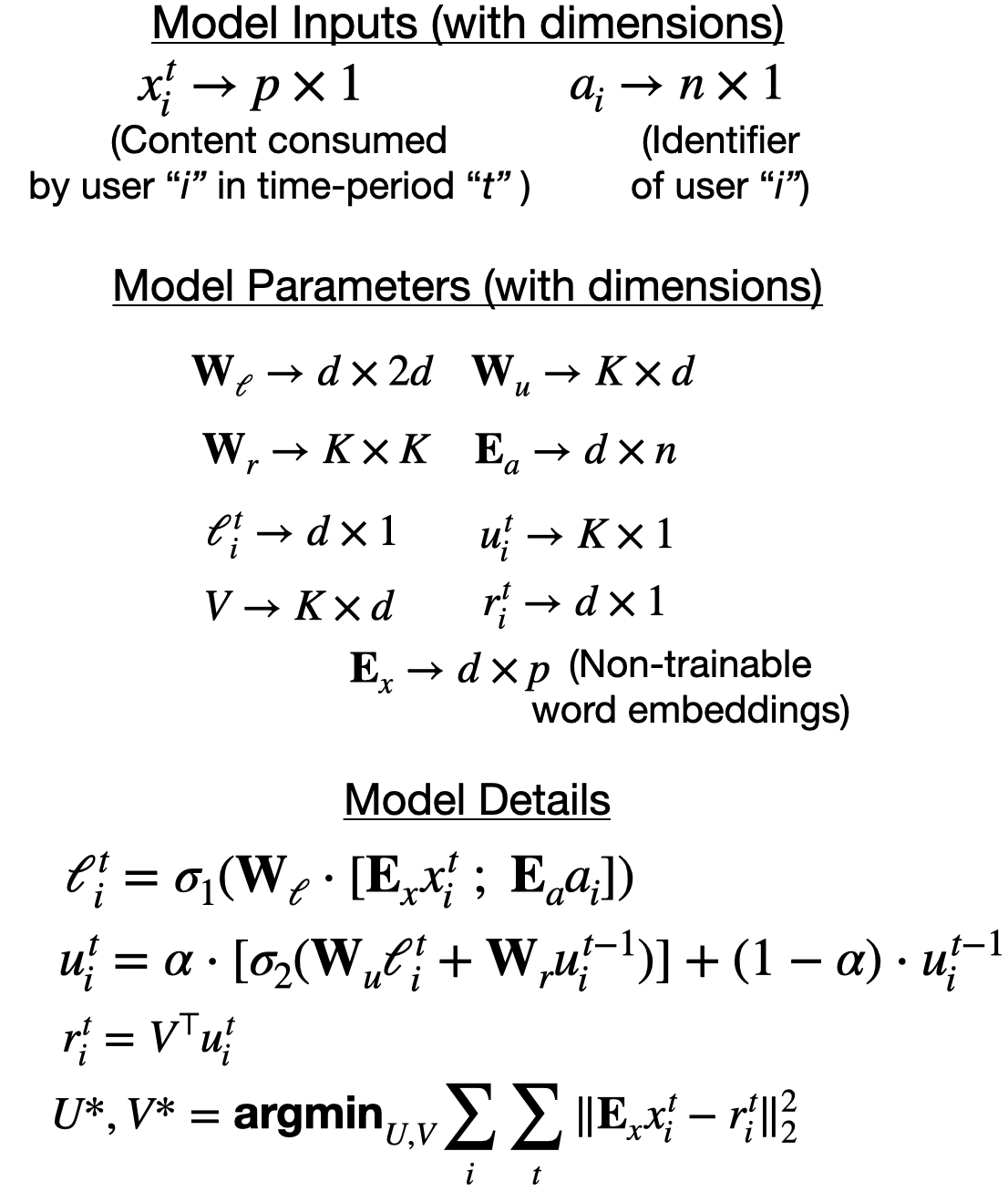} %
   \caption{Neural Net architecture for modeling dynamic user interests.}
   \label{fig-NN-dynamic}
\end{figure}

To summarize, the various details of our model are shown in Figure~\ref{fig-NN-dynamic}.

\subsection{Connection to Other Machine Learning Models}
Our model essentially estimates a dynamic nonlinear low-rank approximation of the input content consumed by users. In the process of doing so, it uncovers latent content attributes as well as each user's evolving tastes over those content attributes. 

Our approach bears connections to several machine learning and natural language processing models that estimate similar low-rank projections for text data. One of the oldest such methods is Latent Semantic Analysis (LSA)~\citep{deerwester1990indexing}. It approximates a document-term matrix, that is, a matrix containing counts of words in each document, with low-rank document and term factor matrices. These estimated factors can then be used, for instance, for information retrieval by computing the similarity between different documents. Our approach is also related to 
Latent Dirichlet Allocation (LDA)~\citep{blei2003latent}, a popular Bayesian generative model of text. The ${\boldsymbol{\beta}}$ and ${\boldsymbol{\theta}}$ topic-word and document-topic probability matrices that LDA estimates can be seen as analogous to our $V$ and {U}$^t$ matrices respectively. A recently proposed probabilistic model, Hierarchical Poisson Factorization (HPF)~\citep{gopalan2015scalable}, also shares some similarities with our model. It also estimates low-dimensional user preference and item attribute factors, though, for modeling implicit feedback data such as movie ratings.

All these models share similarities with our proposed approach. They were proposed in a similar spirit as our model\textemdash to uncover latent low-dimensional structure from high-dimensional text data. However, our approach is different than these methods in modeling, (1) nonlinearities via a novel neural net architecture, (2) dynamics, and (3) incorporating data-driven empirical priors via ``externally estimated'' word embeddings. That said, there are a few probabilistic text models that can model dynamics also, e.g., \citet{blei2006dynamic}, \citet{koren2009collaborative}, and \citet{charlin2015dynamic}. However, their methodological approach is significantly different than ours.

\subsection{Model Estimation \& Optimization}
Neural Net models are estimated just like other statistical models. An estimate of model error (or loss) is computed over the entire dataset. Next, we calculate the gradient of the model parameters with respect to the loss and then move parameters in the direction of the gradient. Due to the nonlinearities, the likelihood function of neural nets, in general, is non-convex. Non-convex objective functions may get stuck in a local minimum or a saddle point and hence can result in getting different parameter estimates based on different parameter initialization. Therefore, one needs to be careful in the optimization of neural network parameters. 

The PyTorch deep learning library was used to estimate our model~\citep{paszke2019pytorch}. We used Adam to optimize our model parameters~\citep{kingma2014adam}, and the learning rate was set at 0.001. The training was performed for 30 epochs when the convergence criteria were met. The model hyperparameters $K$ and $\alpha$ were selected according to the results on a validation set. The  values that were finally selected were $K=30$ and $\alpha=0.5$. The word embedding matrix ${E}_x$ was initialized with pretrained GloVe embeddings. We pretrained the GloVe embeddings (dimensionality $d=300$) on the Globe dataset.\footnote{We also experimented with the pretrained GloVe embeddings downloaded from \url{http://nlp.stanford.edu/data/glove.840B.300d.zip}. The performance of both the sets of embeddings was comparable though the embeddings trained on Globe data were slightly better since we did not need to deal with the issue of out-of-vocabulary words.} The attribute embeddings ${E}_a$ as well as other trainable model parameters were all initialized uniformly at random, as is standard practice. The model estimation was performed on a Nvidia RTX 2080 Ti GPU server with 512 GB of RAM. The model estimation took around 30 hours to converge. Our model has recurrent parts due to temporal dependence between the hidden states. That contributed to the slow model training as it is hard to parallelize recurrent computations.

\section{Results}
This section showcases the empirical performance of our approach in capturing the nuances of users' evolving content consumption tastes.
We divide our results into four parts. First, we visually present the trajectories of users' interests ${U}^t$ as well as the latent content attributes ${V}$ estimated by our model. Then, we perform a crowdsourced study to assess the coherence of the content attributes determined by our model. Next, we turn to quantitative evaluations that highlight the predictive power of the representations learned by our model. Finally, we test the robustness of our findings in several ways, including performing ablation studies to uncover the relative contribution of different aspects of our model.

\subsection{Visualizing Trajectories of User Interests}
\begin{figure}[h]
\begin{center}
\includegraphics[width=0.8\linewidth]{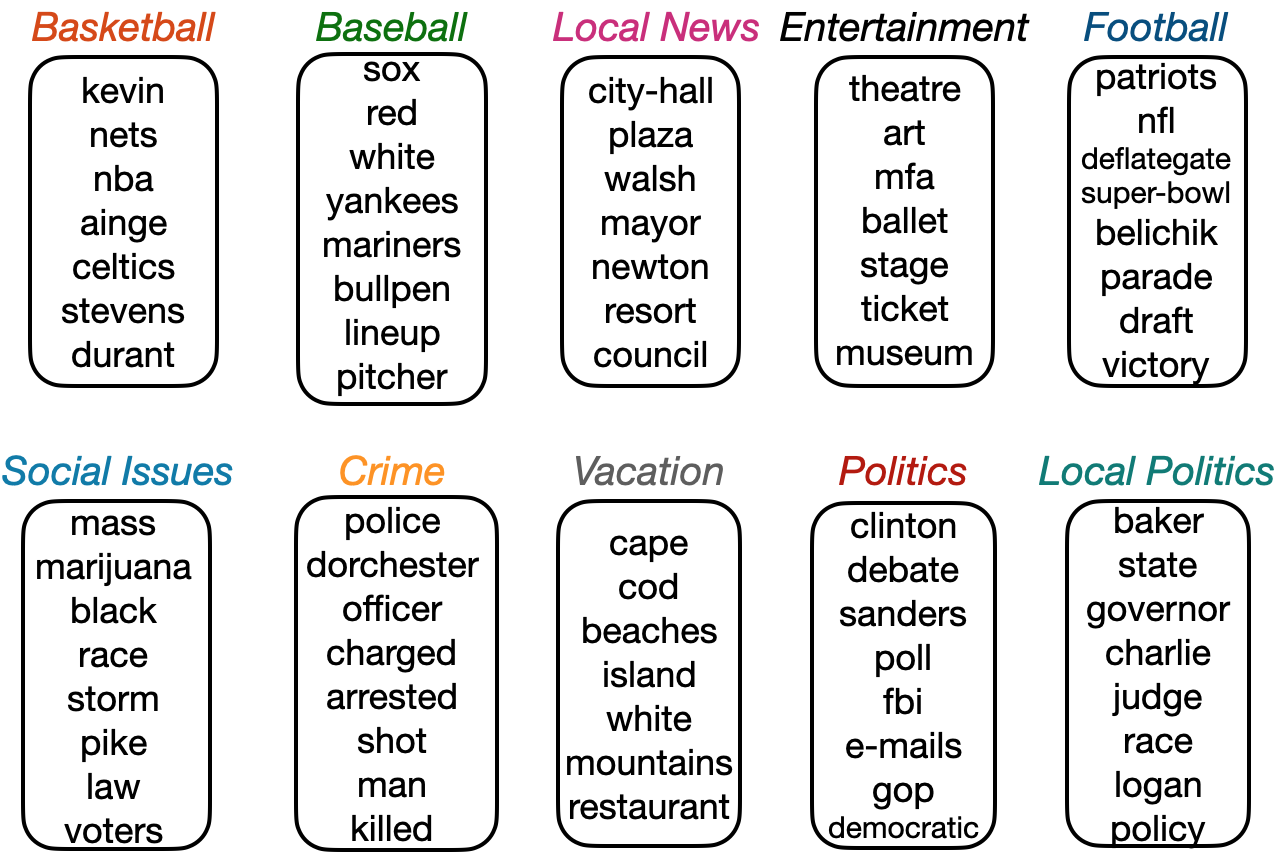}
\caption{The constituent words of a select few content attributes. {\bf Note:} (1) The full list of all the content attributes is in the Online Appendix, (2) Our model outputs a clustered collection of words. The actual names of the content attributes were assigned manually by three research assistants. Whenever there was a conflict, we used the majority label.}
\label{fig:summarize-interests}
\end{center}
\end{figure}

Our model was estimated as described in the previous section. The matrix $V$ uncovers the latent content attributes. For each of its $K$ rows, we found the words associated with that content attribute by computing the nearest neighbors of each row of $V$ from the word embedding matrix ${E}_x$. A set of handpicked content attributes and associated words are shown in Figure~\ref{fig:summarize-interests}.

It is easy to see that the content attributes loosely correspond to intuitive categories of user interests. The topical content of the attributes discovered by our model is more fine-grained than the typical section-based categorization of content by newspaper websites. For instance, several topics relate to sports content, e.g., basketball, baseball, and football, and several that correspond to lifestyles, such as vacation and entertainment. Further, there are some subtle tastes brought to the forefront by our model, e.g., content on social issues, crime, or content related to local (Metro Boston) politics. It is worth emphasizing that the set of content attributes shown in Figure~\ref{fig:summarize-interests} is hand-picked, and like any other mixed-membership text model, our model also results in some less interpretable clusters. For example, one such cluster comprises the words \{house, white, game, thrones, trump, visit, harvard\}, which superimposes politics and entertainment content attributes. The full list of all the content attributes is in the Online Appendix.

As a natural next step, we investigate the evolving tastes of specific segments of users. We focus on three managerially relevant customer personas.\footnote{Based on personal communication with Globe.} Those three customer personas are:\footnote{It is essential to note that, for instance, locals and ex-pats could also consume sports and political content, but it is not what they consume predominantly. Predominant user interests are assumed to account for at least 50\% of their interest weights.}
\begin{itemize}
    \item {\bf Locals and Ex-pats:} These users are mostly interested in local New England news, e.g., local politics, holidays, or sports.
    \item {\bf Sports Fanatics:} As the name suggests these users predominantly consume sports content.
    \item {\bf Political Junkies:} These users mostly consume politics content.
\end{itemize}

For each of these three personas, we classify the corresponding trajectories also into two categories based on temporal trends:
\begin{itemize}
    \item {\bf Stable Interests:} User interests are stable if the relative ranking of their interests does not change over the entire observation period.  
    \item {\bf Evolving Interests:} User interests are assumed to evolve if the relative ranking of their interests changes over the observation period. Further, there are two possibilities. This change in relative rankings of interests could be persistent, or a user could have vacillating interests with fluctuating relative rankings.
\end{itemize}

\begin{figure}[h]
 \centering
    \includegraphics[width=\linewidth]{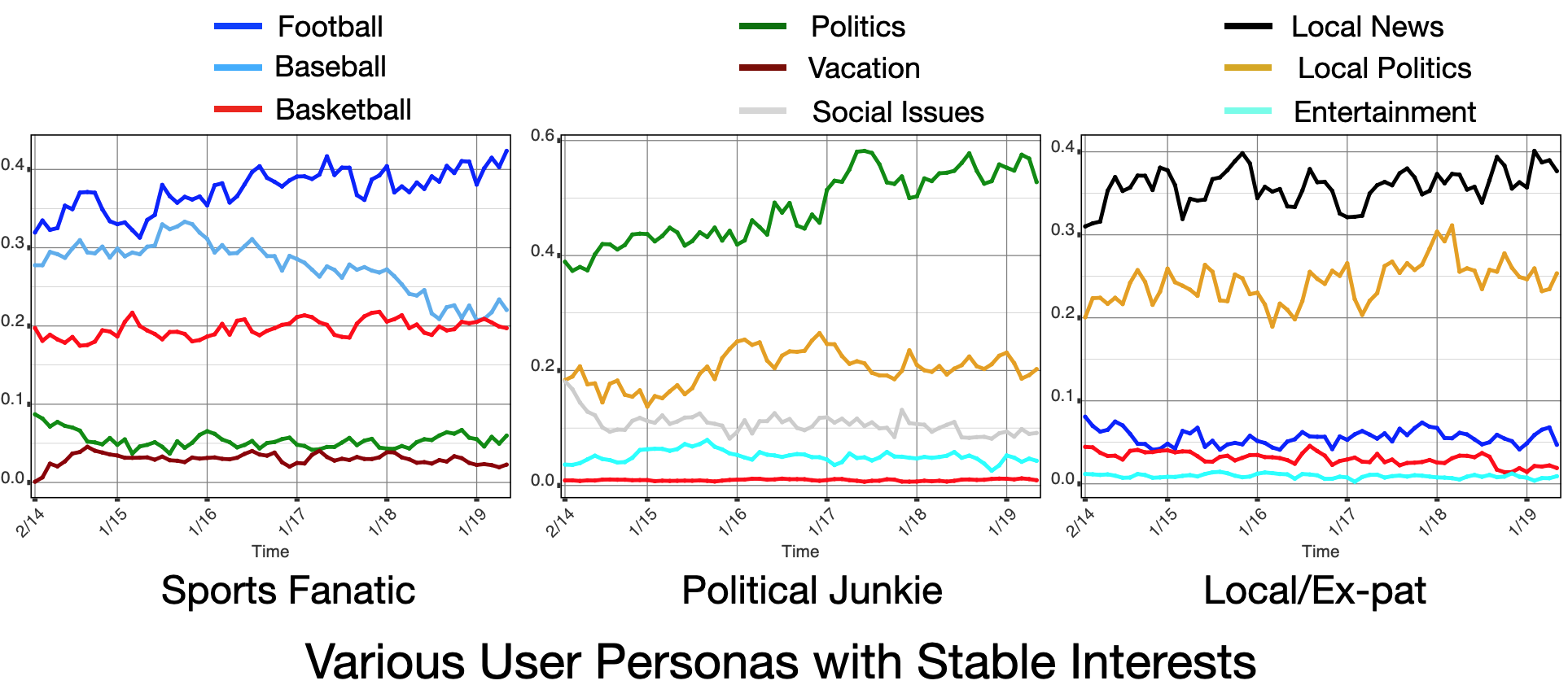}

\caption{(Stable user interests) The plots show the dynamics of the top-5 interests (based on weights from the $U^t$ matrix) of sample users with the different personas. These interests are classified as stable as the relative ranking of these interests does not change. The interests are listed at the top of the figure; the words corresponding to each interest can be found in Figure 5. The y-axis plots the weighting on various interests based on the $U^t$ matrix.}
\label{fig:user_personas}
\end{figure}

Figure~\ref{fig:user_personas} shows the dynamic interests of sample users with the three personas that were just described. These users predominantly consumed content on sports, politics, or local affairs, which can be confirmed from the attribute weights estimated by our model. These were stable user interests as their ranking did not change during our observation period. Our definition of stable user interests concerns the ranking of the interest weights as opposed to the actual weights. For example, the ``Political Junkie'' sample user shown in Figure~\ref{fig:user_personas} had an increase in politics-related content around the time of the 2016 USA Presidential election. However, since that user always consumed high amounts of political content, this increased attention did not impact the rankings of their latent interests, which stayed steady. Similarly, the ``Sports Fanatic'' user had a decrease in their interest in baseball, but they still consumed a high amount of such content relatively. Hence, though, the actual interest weights could shift over time, but that even might not indicate a significant departure from the status quo as the relative rankings are stable. 

\begin{figure}[h]
 \centering
    \includegraphics[width=\linewidth]{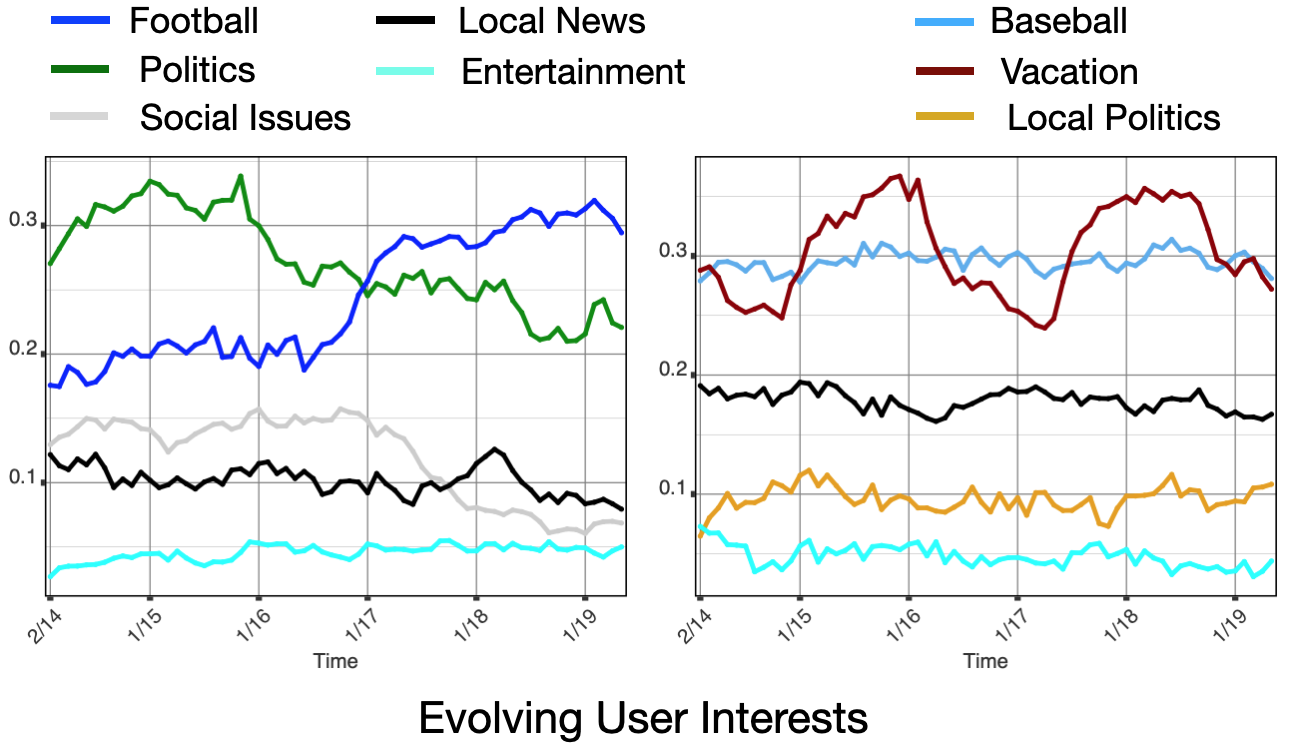}

\caption{(Evolving user interests) The plots show the dynamics of the top-5 interests (based on weights from the $U^t$ matrix) of sample users with evolving interests. The interests are listed at the top of the figure; the words corresponding to each interest can be found in Figure 5. The y-axis plots the weighting on various interests based on the $U^t$ matrix.}
\label{fig:evolving_interests}
\end{figure}

Along similar lines, Figure~\ref{fig:evolving_interests} shows two users with evolving interests. As opposed to stable interests, we assume users have evolving interests when there is a change in the ranking of their interests. This change can further occur in two different ways. There could be a persistent change in the rankings, which could potentially be due to a permanent change in the underlying content preferences. The left panel in Figure~\ref{fig:evolving_interests} shows a user with persistently evolved interests. Starting around January 2017, their interest in politics waned, and they started paying more attention to football news.

The user interests could also vacillate, leading to a temporary change in rankings. The right panel in Figure~\ref{fig:evolving_interests} shows such a user. As can be seen, the sample user's interest in vacation-related content waxes and wanes over time, perhaps due to their seasonal interest in such content or due to a fluctuation in their underlying preferences. Needless to say, but this distinction between stable and evolving interests is a valuable piece of information for a marketing analyst who is monitoring this user and wants to intervene.

To summarize, our approach uncovers both evident and nuanced trends in user interests. The classification of interests into stable and evolving captures a critical distinction in the underlying user preferences and can be leveraged by marketers to tailor their messages to the user. Depending on the context, evolving user interests could, for instance, indicate purchase intent, or they might suggest the need for a personalized nudge.\footnote{The stratification of user interest trajectories into stable and evolving was developed by us to summarize managerially meaningful dynamics of user interests.}

\subsection{Crowdsourced Evaluation of Content Attributes}
To further solidify the qualitative evidence presented by the trajectories of user interests, we perform a crowdsourced evaluation. The trajectories that we visualized appear to capture nuanced user tastes, but lack impartial human assessment. So, we use Amazon Mechanical Turk (MTurk) to perform a human evaluation of the coherence of the content attributes estimated by our model.

To assess the efficacy of the content attribute matrix $V$ in capturing coherent and meaningful concepts, we perform the word intrusion task as specified by \citet{chang2009reading}. In the word intrusion task, human subjects have to identify the {\it intruder} words from the list of words belonging to a topic or a content attribute in our case. For example, in the list of words \{celtics, bruins, canadiens, rangers, apple, giants\}, most people identify {\it apple} as the intruder word since the other words make sense together (they are names of sports teams).

We follow the evaluation strategy outlined by \citet{chang2009reading} firmly. For each content attribute represented by a row of the $V$ matrix, we choose five words that are closest to it in terms of cosine-similarity. Next, we choose an intruder word that has a lower similarity to a given row of the $K$ matrix but has a higher similarity to another row of $K$. Finally, all six words are randomly shuffled and presented to human subjects. The human judgments are evaluated using the model precision metric defined in Equation~\ref{eq:precision}, where $k$ indexes the content attributes (the row index of $V$), and $S$ is the total number of human subjects. The variable $i_{k,s}$ denotes the intruder word defined by a human subject for a particular content attribute; $w_k$ is the correct intruder word that was used for that content attribute. 

\begin{equation}
    \text{Mean Precision}_k=\sum_{s=1}^S \frac{{\bf 1}(i_{k,s}=w_k)}{S} 
    \label{eq:precision}
\end{equation}

\begin{figure}[htbp]
    \centering
    \includegraphics[width=0.5\linewidth]{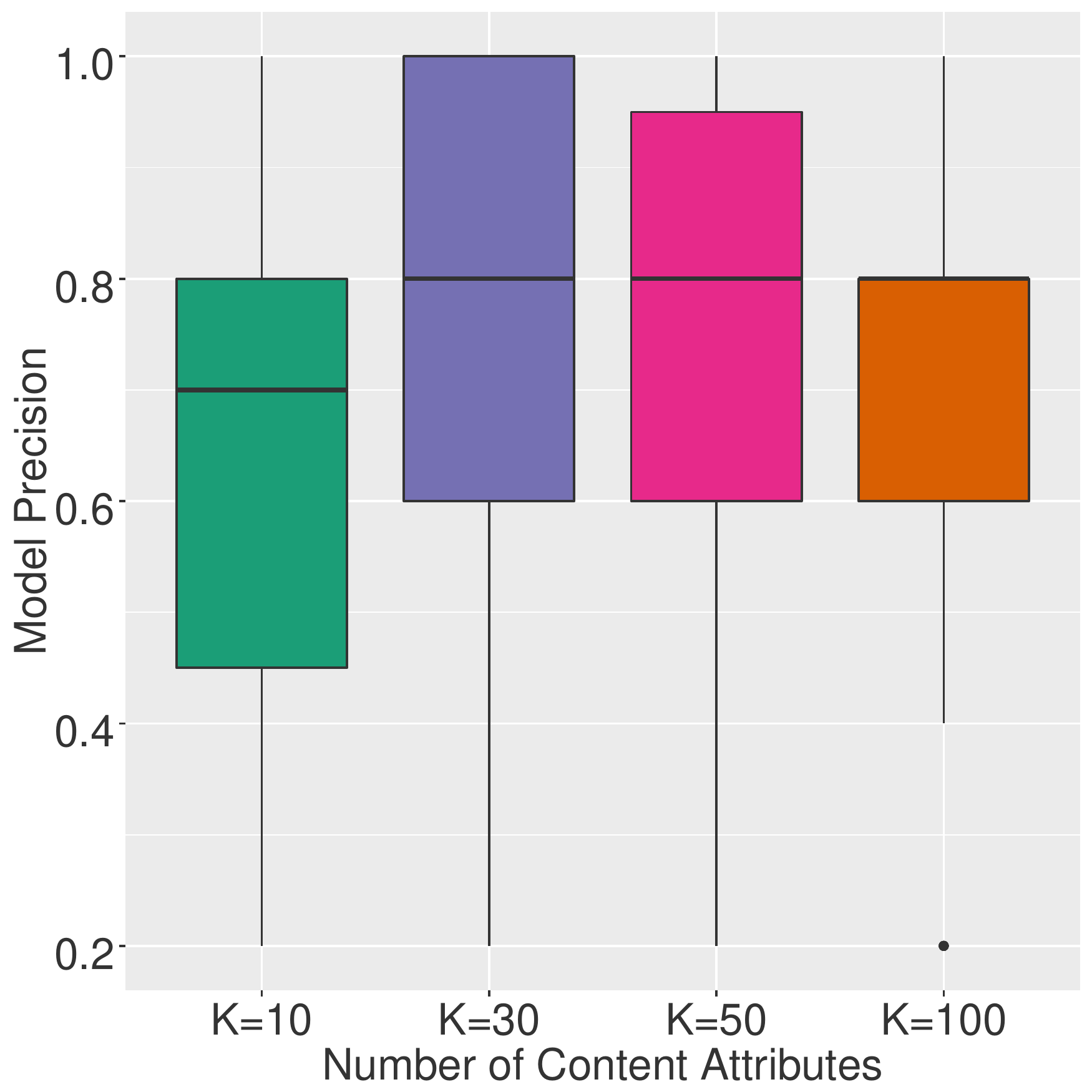}
    \caption{Crowdsourced Mean Model Precision for different number of attributes.}
    \label{fig:boxplot}
\end{figure}

We collected judgments from five different Amazon Mechanical Turk (MTurk) workers. We asked each worker to detect the intruder word for each content attribute, that is, a word intrusion task for each row of $K$. Since this is a qualitative analysis of the coherence of the content attributes estimated by our model, we evaluated our model with four different dimensionalities of the attribute matrix, $K=\{10, 30, 50, 100\}$.  Mean model precision will be one if all the five workers can find the correct intruder word and zero if all of them selected the wrong intruder word. A higher model precision indicates greater coherence of a content attribute since a higher number of human judges were able to spot the intruder word easily. The box plot in Figure~\ref{fig:boxplot} shows the model precision for our model for various values of $K$. As the boxplot suggests, our content attributes generally exhibit high model precision and hence a high-degree of cohesiveness.

\subsection{Evaluating the Predictive Quality of the Estimated Dynamic User Interests.}
Our model generates trajectories of user interests based on the content they consume. As we just saw, they are coherent and unravel nuanced user behaviors in content consumption. Next, we build on those results by showcasing the predictive power of these learned user representations. The user and content factors estimated by our model together provide a low dimensional summarization of a user's content consumption profile. Hence, we use the d-dimensional reconstruction vector $r_{i}^{t}(=V^{\top}u_i^{t})$ output by our model to assess the predictive power of the representations learned by our model. In particular, we use $r_{i}^{t}$ estimated till $t=\tau-a$, that is, up till $a$ previous time periods in our observation period to predict a user's content consumption in the final $\tau^{th}$ time period. The time-period $\tau$ may not be aligned across calendar time for all the users as it is the last time a user consumed content. Hence, each user has a potentially different value of $\tau$ in calendar time. We chose not to use a subscript $\tau_i$ as that adds further complexity to our notation.

Predicting what a user will read next seems like a cumbersome task from a statistical modeling perspective as the output is a high-dimensional text vector. There are thousands of unique news articles that can be read by a user, which makes it a classification problem with thousands of output classes. Given the nature of news articles, most of these output classes appear only a few times in the dataset. Hence, we need to simplify this prediction task into one that can be solved easily.

We construct two empirical tasks to highlight the superior predictive quality of the user representations learned by our model. At the heart of both these prediction tasks are the d-dimensional vectors $r_{i}^{\tau-a}$ and $c_{i}^{\tau}$. Just to recall, $r_{i}^{\tau-a}$ is the user factor estimated using data from the first $\tau-a$ time-periods. Since $r_i^{\tau-a}$ lies in the same d-dimensional space as the input, it is also called the reconstruction vector. Whereas, $c_{i}^{\tau}$ is the object that we want to predict, that is, the d-dimensional embedding of the content consumed by the user in the $\tau^{th}$ time-period. We generate the embeddings for $c_{i}^{\tau}$ using the pretrained GloVe embeddings via the same procedure as described in Section 4.3.1 for input embeddings.

The user factor $r^{\tau-a}_i$ can be represented as a point in a d-dimensional space. If it indeed captures the subtle patterns in a user's dynamic interests, then one should expect it to be proximate to the d-dimensional content embedding $c_{i}^{\tau}$. We use this intuition to guide our evaluation strategy. As part of our first evaluation, we find the nearest neighbor (in terms of the cosine similarity) of each user's representation vector $r^{\tau-a}_i$ from the content embeddings. The mean precision is computed as the fraction of users for whom the nearest neighbor was their own content embedding $c_{i}^{\tau}$. More precisely, our evaluation metric is,

\begin{equation}
    \text{Mean Precision}= \sum_{i=1}^n \frac{{\bf 1}(\text{NN}_1(r_i^{\tau-a})=c_{i}^{\tau})}{n}
    \label{eq:NNaccuracy}
\end{equation}
where, $\text{NN}_1(\cdot)$ represents the nearest neighbor function that returns the content embedding that is closest in terms of cosine similarity to the user factor. $n$ is the total number of users in our evaluation set. This metric is also known as ``mean precision at K (MP@K)'' in the information retrieval literature.\footnote{\url{https://en.wikipedia.org/wiki/Evaluation_measures_(information_retrieval)}} In our case, we only consider one nearest neighbor, so essentially we are calculating ``MP@1''. We also evaluated mean precision for more nearest neighbors, in particular MP@3, MP@5, and MP@10, and the trends in results were remarkably similar. Though, the actual mean precision was higher as the retrieval problem becomes easier with an increase in the number of nearest neighbors considered.

Our second evaluation builds on the first one and captures the proximity of embeddings on a continuous scale instead of an all-or-nothing nearest neighbor prediction. So, we compute the real-valued similarity score $s(r_{i}^{\tau-a},c_{i}^{\tau})$ between the d-dimensional user and content vectors. More precisely, we compute the cosine similarity between the vectors. Once again, we draw on the ability of high-quality representations to cluster together in the d-dimensional embedding space.

We split our dataset into two parts\textemdash training and validation. The data is shuffled randomly, and the training/validation splits are constructed with 90\%/10\% of users, respectively. We estimate Equation 9 on the training dataset and then tune the model hyperparameters $K$, $\alpha$ on the validation dataset. The details of hyperparameter tuning are described in the next subsection. Since our evaluation involves computing the nearest neighbors and similarity of embeddings, which do not have any estimable parameters of their own, we do not need a separate held-out test set. Hence, we use the training data itself for the nearest neighbor retrieval and similarity tasks. All our models are estimated on the first $\tau-a$ time-periods. We only access the content consumption in $\tau^{th}$ time-period while benchmarking the prediction (or retrieval) accuracy of the learned representations.

We benchmark the predictive quality of the representations learned by our model by comparing its performance against several alternatives. Three out of the four options that we consider broadly fall into the class of ``topic models.'' At a high level, they posit a data generating process that assumes the text is generated by several underlying latent factors called topics. The fourth baseline that we compare against is a weighted average of the content consumed by a user across different newspaper website sections.   

\begin{enumerate}
\item {\bf Latent Dirichlet Allocation (LDA):} LDA is a popular hierarchical Bayesian model of text generation~\citep{blei2003latent}, which has been used in several marketing analytics applications~\citep{buschken2016sentence,liu2017consumers}. It describes a data generating process for collections of text data such as documents where each document contains a set of words. It assumes that a small number of latent topics generate each document. And, each word is further created by one of these topics.

LDA was not proposed for modeling dynamic user interests, which is the problem that interests us. However, it can be adapted to model user interests by assuming that the total content consumed by each user $\{x_i^t\}^{t=1:\tau}$ is a document. Then, the topic-word matrix estimated by LDA $\beta$ is analogous to our matrix $V$, and the document-topic matrix $\theta$ is comparable to the matrix $U^{t}$. For an apples-to-apples comparison with our approach, we need the equivalent of our d-dimensional user factor $r_i^{\tau-a}$. Once we have that, then we can easily compute the nearest neighbor and the similarity score.  

It is rather straightforward to generate the equivalent of $r_i^{\tau-a}$ for LDA. We use the $\beta$ matrix to find the top 50 words which have the highest posterior probability for each topic and then extract their pretrained GloVe embeddings. Next these embeddings are averaged over  all the words in a given topic, thereby generating a d-dimensional vector for each of the $K$ topics. Finally, we multiply these embeddings with the document-topic matrix $\theta$ to output a d-dimensional user factor similar to $r_i^{\tau-a}$. 

We tried $K=\{30,50,100,200\}$ for the number of LDA topics and chose the value that gave the best accuracy on the validation dataset. We got the best performance at $K=50$. We train LDA for 100 iterations with a collapsed Gibbs sampler. To make as close a comparison as possible, LDA is also trained on the content consumed by each user in the first $\tau-a$ periods only.

\item {\bf Dynamic Topic Model (DTM):} DTM~\citep{blei2006dynamic} is the dynamic version of LDA. It assumes that the topic mixtures per document remain the same over time, but topics themselves evolve. In comparison to our model, it assumes that a user's weighting over the content attributes $U$ is static, but the content attributes $V$ themselves drift over time. 

We adopt a similar procedure as defined above for LDA to generate a d-dimensional user factor for DTM.  We tried $K=\{30,50,100,200\}$ for the number of DTM topics and finally chose $K=30$ as it gave the best accuracy on the validation dataset. The rest of the estimation procedure for DTM exactly mirrors that of LDA.

\item {\bf LDA-Gaussian Process Dynamic Heterogeneity (LDA-GPDH):}
Next, we compare our approach against LDA-GPDH~\citep{dew2017dynamic}, which is a flexible approach for modeling dynamic heterogeneity using Gaussian Processes. It is proposed for modeling the evolution of product reviews but can be easily adapted to model dynamic user interests. 

Unlike DTM, LDA-GPDH assumes that the topics are static, but the mixture of topics per document changes over time. A Gaussian Process parameterizes the fluctuation of a topic from its mean prevalence in a document. So, similar to our model, LDA-GPDH assumes that a user's weightings over the different topics evolve, but the topics themselves remain static. The parameters $\nu_d$ and $\beta_{id}(t)$ as presented in \citet{dew2017dynamic}, where d indexes the topics and $i$ indexes the products, correspond to our matrices $V$ and $U^t$ respectively. Similar to LDA and DTM, we map the topic-word probability distribution of LDA-GPDH to d-dimensional GloVe embeddings and generate a user factor corresponding to our $r_i^{\tau-a}$.

The rest of the estimation and evaluation procedure is similar to that of LDA and DTM. We tried K=$\{15,30,50,100\}$ number of topics and got the best validation accuracy for $K=15$.\footnote{LDA-GPDH was estimated using the code provided by \citet{dew2017dynamic} in personal communication.}

\item {\bf Weighted Average of Topical Content:} Globe categorizes content on its website into sections, e.g., politics, sports, metro, opinion, business, etc. These content categorizations are generated manually by the editorial team. So, a natural baseline for predicting a user's future content consumption is the weighted average of content consumed by them in the past. To be more precise, we compute the average GloVe embeddings of the 50 most frequent words that a user consumed from various sections and then weight those embeddings by the overall share of content consumed by the user from each of those sections. We compute these weighted content embeddings using content from the first $\tau-a$ time periods to generate the equivalent of $r_{i}^{\tau-a}$. The rest of the estimation and evaluation procedure is the same as for LDA, DTM, and LDA-GPDH.

\end{enumerate}

Tables~\ref{tab:content_predict_1nn} and~\ref{tab:content_predict_cos} benchmark the performance of various models. The results show the superior performance of our approach with striking consistency across different time-horizons of prediction (a=1, 2, 3) for both the nearest neighbor retrieval precision and cosine similarity evaluation metrics.

A bit unsurprisingly, the baseline model, which uses a weighted average of the existing sections on the Globe website, performed the worst. It suggests the benefits of a data-driven categorization of news stories in being predictive of latent user interests. The two dynamic models DTM and LDA-GPDH, were the most competitive baselines, though they still performed significantly worse than our model. The two critical dimensions along which our model differs from these baselines are in modeling nonlinearities via a neural network and in performing exponential smoothing of the user trajectories. The strong performance of our model corroborates similar findings by the deep learning community~\citep{goodfellow2016deep} of the superiority of neural networks in extracting nonlinear patterns from large datasets. Also, since most users' trajectories are relatively short (Figure~\ref{fig:user_activity}), exponential smoothing improves predictive accuracy by acting as a regularizer.

\begin{table}[ht]
   \centering
     \begin{tabular}{l|c|c|c} 
      \hline
      &\multicolumn{1}{c}{\bfseries $a=1$}&\multicolumn{1}{c}{\bfseries $a=2$}&\multicolumn{1}{c}{\bfseries $a=3$}\\
      \cline{2-4}
         Method& Mean Precision  &Mean Precision & Mean Precision \\
      \hline
      \hline
       {Weighted Average of Sections}& 3.8&2.2 &1.4 \\
       {LDA} & 10.4 &7.8&6.4\\
       {LDA-GPDH}& 12.2 &10.7&8.7 \\
       {DTM}& 14.9 &12.6&10.9  \\
      {\bf Our approach}& {\bf 17.1  }& {\bf 15.6 }& {\bf 13.2} \\
      \hline
      \hline
   \end{tabular}
   \caption{Results on the task of retrieving the nearest neighbor, i.e., MP@1. The models are estimated on data up till $a$ previous time periods. The prediction is always made on content consumption in the final $\tau^{th}$ period. {\it Note: } 1) Mean Precision represents the fraction of users whose nearest neighbor was retrieved correctly. Please refer to Equation~\ref{eq:NNaccuracy}, 2) Precision numbers are multiplied by 100 to standardize them, 3) Table shows training set accuracy, 4) Model hyperparameters were tuned on the validation dataset.}
   \label{tab:content_predict_1nn}
 \end{table}
 
 \begin{table}[ht]
   \centering
     \begin{tabular}{l|c|c|c} 
      \hline
      &\multicolumn{1}{c}{\bfseries $a=1$}&\multicolumn{1}{c}{\bfseries $a=2$}&\multicolumn{1}{c}{\bfseries $a=3$}\\
      \cline{2-4}
         Method& Similarity ($\mu\pm\sigma$) & Similarity ($\mu\pm\sigma$)& Similarity ($\mu\pm\sigma$) \\
      \hline
      \hline
       {Weighted Average of Sections}& 42.9 $\pm$ 10.6&40.1 $\pm$ 9.4 &38.7 $\pm$ 9.2 \\
       {LDA} & 55.4 $\pm$ 5.1 &52.9 $\pm$ 4.6&50.1 $\pm$ 5.4 \\
       {DTM}& 64.6 $\pm$ 2.1 &61.2 $\pm$ 2.9&58.6 $\pm$ 4.3  \\
       {LDA-GPDH}& 62.8 $\pm$ 3.0 &61.0 $\pm$ 2.4&59.9 $\pm$ 4.0 \\
      {\bf Our approach}& {\bf 71.3 $\pm$ 3.3 }& {\bf 69.4 $\pm$ 3.9}& {\bf 67.0 $\pm$ 3.6 } \\
      \hline
      \hline
   \end{tabular}
   \caption{Resuls showing cosine similarity between embeddings of users and the content they consumed. The models are estimated on data up till $a$ previous time periods. The prediction is always made on content consumption in the final $\tau^{th}$ period. {\it Note: } 1) Similarity represents the cosine similarity $\frac{a\cdot b}{\|a\|\|b\|}$, 2) Similarity numbers are multiplied by 100 to standardize them, 3) Table shows training set accuracy, 4) Model hyperparameters were tuned on the validation dataset.}
   \label{tab:content_predict_cos}
 \end{table}

Our model unpacks users' complex content consumption patterns by estimating an interpretable dynamic probabilistic weighting over a set of key underlying interests. Further, the user representations learned by our model embed closer to their future content consumption embedding and hence wield predictive power. Thus, a firm can use our results to recommend specific news articles or broad content topics to the users. In its simplest form, such a recommendation can be made by computing the cosine similarity between the reconstruction vector $r_i^{\tau-a}$ and the candidate news stories published on a given day $c_i^{\tau}$ and then recommending the top few items. Alternatively, one can choose items to recommend based on the nearest neighbors of the user representations $r_i^{\tau-a}$. Such conceptualizations formed the basis of some of the earliest deployed recommender systems~\citep{sarwar2002recommender,koren2009matrix}. This was partly the reason that we designed our predictive evaluation based on these metrics.

\subsection{Other Important Analyses: Robustness Tests, Ablation Analyses, and Real-World Deployment Challenges}
Our model makes several design choices, including the selection of tunable hyperparameters. Further, several essential modeling details are crucial to get right for the successful deployment of our model. So, as a next step, we test the sensitivity of the model performance to these design choices and explain the key engineering details to aid the scalable deployment of our model. We divide our analysis into three parts.
\subsubsection{Robustness Tests}
We check the robustness of our model to two different hyperparameter choices. We consider several choices for the number of content attributes $K= \{10,30,50,100\}$ and the amount of exponential smoothing $\alpha= \{0.10, 0.25,0.50,0.75,0.90\}$. We train our model on 90\% of the data and compute the nearest neighbor accuracy on the validation data (10\%). Finally, the best performing hyperparameters were chosen. The results are shown in Table~\ref{tab:hyper}. As can be seen, averaging over different values of $\alpha$, the best value of $K$ was 30, and the best value of $\alpha$ was 0.5 while averaging over different values of $K$. These hyperparameter values also provided the best held-out accuracy when used together.

\begin{table}[ht]
   \centering
     \begin{tabular}{l|c|c|c|c|c} 
      \hline
       &\multicolumn{5}{c}{Mean Nearest Neighbor Precision} \\
      \cline{2-6}
         Hyperparameters&$\alpha=0.10$&$\alpha=0.25$ & $\alpha=0.50$& $\alpha=0.75$&$\alpha=0.90$\\
      \hline
      \hline
       {K=10}&12.9&14.1&15.2&15.0&13.6\\
       {K=30} &12.1&15.8&{\bf 18.4}&16.6&14.4\\
       {K=50}&11.2&15.3&17.7&15.2&14.1 \\
       {K=100}&13.4&16.2&17.5&16.8&14.5\\
      \hline
      \hline
   \end{tabular}
   \caption{Table showing the impact of hyperparameter choice on the validation set accuracy. {\it Note: }1) Mean Precision represents the fraction of users whose nearest neighbor was predicted correctly. Please refer to Equation~\ref{eq:NNaccuracy}, 2) Precision numbers are multiplied by 100 to standardize them.}
   \label{tab:hyper}
 \end{table}

\subsubsection{Ablation Analysis}
Next, we perform several ablation analyses by unraveling various components of our model. Essentially, we ``turn off'' certain parts of our model and evaluate the predictive ability of the rest of the model. These ablation studies allow us to quantify the relative contribution of the multiple design choices in our model. 
\begin{itemize}
\item {\bf The contribution of nonlinearities:}
Our model incorporates nonlinearities that are parameterized by a neural network. So, it is natural to benchmark our model's performance against a model that does not contain any nonlinearities. Hence, we take our model as described in Figure~\ref{fig-NN-dynamic} and remove all nonlinearities such as the activation functions $\sigma_1$, $\sigma_2$ and the associated parameters \{${W}_{\ell}$, ${W}_u$, and ${W}_r$\}. We keep everything else the same, including the loss function and gradient-based model training via the Adam optimizer. This modified model is then used to estimate the predictive quality of our dynamic user representations via the nearest neighbor prediction task. Results shown in Table~\ref{tab:ablation} illustrate the contribution of nonlinearities towards the model performance. As can be seen, a model with no nonlinearities significantly underperforms the full model.

\item {\bf The impact of modeling time dynamics:}
Modeling the time dynamics of users' content consumption is central to our model. So, an interesting counterfactual to consider is the case when there is no time dimension in our model. This scenario can be simulated by assuming that the variable $x_i^t$ (Figure~\ref{fig-NN-dynamic}) contains the content consumed by each user over the entire observation period. In our actual model, however, the variable $x_i^t$ contains only the content consumed during the time-period $t$. All other details of our model remain the same as earlier.

Again, we use this model with no time dimension to make nearest neighbor prediction. The resulting accuracy of the model is shown in Table~\ref{tab:ablation}. The results show a significant decrease in accuracy in the absence of modeling time dynamics, thereby underscoring the importance of modeling the time dimension and the drift of users' content consumption tastes.
    
\item {\bf The impact of exponential smoothing:}
Finally, we quantify the impact of exponential smoothing in our model. As described earlier, we perform exponential smoothing to ensure that the trajectories of users' interests evolve smoothly over time. In other words, exponential smoothing can be seen as providing valuable regularization to our model, which improves its generalization performance. We remove the exponential smoothing from our model by setting $\alpha=1$ in Equation~\ref{eq:exp_smooths}. The accuracy of the resulting model dropped substantially once again, as can be seen in Table~\ref{tab:ablation}.
\end{itemize}

\begin{table}[ht]
   \centering
     \begin{tabular}{l|c|c|c} 
      \hline
       &\multicolumn{3}{c}{Mean Nearest Neighbor Precision}\\
      \cline{2-4}
         Ablation&a=1 & a=2&a=3 \\
      \hline
      \hline
       No nonlinearities&11.7&8.6&6.5\\
       No Time Dynamics &13.1&10.3&8.1\\
       No Exponential Smoothing&15.7&12.9&10.8 \\
       \hline
       Full Model (Table~\ref{tab:content_predict_1nn})&{\bf 17.1}& {\bf 15.6}&{\bf 13.2} \\
      \hline
      \hline
   \end{tabular}
   \caption{Table showing the relative contribution of various components of our model. Training set accuracy is reported. The models are estimated on data up till $a$ previous time periods. {\it Note: }1) Mean Precision represents the fraction of users whose nearest neighbor was predicted correctly. Please refer to Equation~\ref{eq:NNaccuracy}, 2) Precision numbers are multiplied by 100 to standardize them.}
   \label{tab:ablation}
 \end{table}

The various ablation studies paint a coherent picture of the importance of modeling nonlinearities, time dynamics, and performing exponential smoothing on model performance. Excluding any of these components leads to a substantial decrease in model accuracy. Among the various model parts, nonlinearities and time dynamics seem to be the most crucial elements in terms of providing superior model performance.

\subsubsection{Real-World Deployment Challenges: Scalability, Transferrability, and Cold-Starting New Users}

There are several related challenges involved in the real-world deployment of our model. The key underlying issue is regarding dealing with the arrival of new users. More precisely, how can we use our estimated model to generate the consumption trajectories $u_i^t$ for new users? This can be further divided into two parts. Do we have consumption traces $x_i^t$ for these users, or are they first-time visitors?

\begin{itemize}
    \item {\bf Use Transfer Learning to learn representations for users with consumption traces $x_i^t$:}  
     This challenge arises in two real-world scenarios faced by any digital marketing analytics firm. First, the model has been estimated on a fixed set of users, and new users arrive. The firm has access to the content they consumed $x_i^t$; however, estimating the model every time there is an influx of new users is impractical. Can we use an already estimated model to induce the representations for these new users? The second challenge arises due to computational concerns. The firm has estimated our model on a small but representative subpopulation of users, so can we transfer the representations learned by this model to the full population of users? This scenario is also encountered while estimating the model for generating the results described in this paper. There are recurrent components in our Neural Net, which makes it hard to parallelize. Hence, we estimated our model on a random subsample of 500,000 users, and we would like to scale it to our full user-base.

    It turns out that there is a simple and efficient solution to this Transfer Learning problem. Recall that the key parameters estimated by our model are $\{U\}^{t=1:\tau}$ and ${V}$, as also shown in Equation 9. Of these parameters, the content factors ${V}$ are shared by all the users and can be thought of as learning a common ``basis'' to represent the content. Hence, as long as the initial user subpopulation used to estimate the model is representative, we can use the estimated ${V}$ for new users. So, for a new user $s$, we only need to estimate their dynamic weighting over the content factors, that is, $u_s^t$. This can be done easily by freezing all the estimated parameters \{${ W}_{\ell}$, ${W}_{u}$, ${W}_{r}$, ${V}$, ${E}_x$\} of our model as shown in Figure 3, except ${E}_a$ (initialized randomly) and then feeding it the consumption traces $x_s^t$, and a user identifier $a_s$. Once again, we estimate this model using Adam via backpropagation, with the only estimable parameters being the d-dimensional user embeddings contained in the matrix ${E}_a$. Once the model estimation has finished, we have an estimate of the new user's dynamic weighting $u_s^t$ as desired. This trivial estimation can be performed very fast, unlike the full model training, since we estimate only one set of parameters while fixing all others. It has the effect of making the optimization landscape less non-convex than learning all the parameters at once. Hence, we can transfer the representations learned by our model to new users by incurring only a small computational cost.

    \item {\bf Cold-starting new users with no consumption traces:} 
    Any successful real-world deployment of the model would also need to cold-start the new users, that is, learn representations for whom we have no observed content consumption traces. In this realistic but even more challenging scenario, we can not use Transfer Learning as described above to estimate new user representations. Instead, the only recourse in this scenario is to use the observable user demographics, such as zip code, desktop/mobile, age, etc., to find the nearest neighbors of the new users from among the users for whom we have already estimated the model. Finally, the estimated dynamic weighting for the new users can be a simple average of their neighbors' weightings, which can be used to generate an initial set of item recommendations. Note that throughout this paper, we have never used the observable user demographics before, but any digital marketing company has access to them. And, they will come in handy in generating cold-start recommendations.
\end{itemize}

\section{Discussion}

This paper proposed a neural matrix factorization method to extract nonlinear patterns from high-dimensional text data. We used it to model the dynamics of users' content consumption interests. Our results highlight the superior ability of our model in capturing nuances in dynamic consumption patterns. Each user's estimated interests open a window into their evolving tastes and can be used to create data-driven user personas that are predictive of their future content engagement. These personas or the embeddings themselves $r^t_i$ can be used, for instance, to build user profiles, to recommend news articles, or to create personalized news categorizations with a few caveats. In addition to neatly summarizing user interests, the estimated low-dimensional user profiles also have high predictive power. Our approach significantly outperforms a host of competitive baseline methods in predicting future user engagement.

Methodologically our model represents significant advances over existing approaches. The dynamic matrix factorization formulation of our method allows us to decompose a user's news consumption into a set of latent content attributes coupled with that user's dynamic weighting over those attributes. Such a natural decomposition of a customer's journey aids with the interpretability of our findings. Further, our neural net model combines with this simple matrix decomposition to help us model flexible nonlinear dependence between the high-dimensional textual content and users' latent interests. Hence, our method provides the best of both worlds. It combines the benefits of flexible nonlinear neural net modeling and the simplistic interpretation of matrix factorization. Our approach permits this while also seamlessly incorporating temporal dependence between user interests. Finally, the ability to incorporate empirical data-driven priors into our model in the form of pretrained word embeddings estimated on external data sources provides a significant comparative advantage to our model.

 To the best of our knowledge, this is the first paper to propose a novel neural net architecture for a relevant marketing problem. Extracting patterns from high-dimensional text data is a common problem faced by digital marketers these days. Our paper is also the first paper to apply matrix factorization style methods to a digital marketing problem while also highlighting the simplicity of such methods.

\subsection{Managerial Implications} 
Our model provides an end-to-end customer analytics framework that can be used by marketing managers to profile the users, track the health of their customer-base, and design suitable interventions for retaining them. To that end, our results have important managerial implications.

\paragraph{Generating User Profiles:} The trajectories of latent user interests estimated by our model provide a concise summary of their often fluctuating and evolving underlying content preferences. Since these trajectories provide a dynamic weighting over a set of underlying content attributes, they are also easy to interpret. Further, these user representations are estimated from fine-granular user interactions with the news content. These considerations make these representations a perfect candidate for building user profiles. A user profile summarizes a user's interests revealed via their behavioral patterns online and has numerous digital marketing applications, including the targeting of advertisements.  The computational efficiency and ease of estimation of our model, coupled with its ability to harness highly predictive subtle dynamic cues from large datasets, would make it an excellent choice for industrial deployment.

Above and beyond the utility of user-profiles in digital marketing applications, marketing managers can also use the trajectories generated by our model for an initial sniff test to detect anomalous patterns in individual-level consumption behavior. Any idiosyncratic deviations could be used to trigger a personalized intervention, for instance.

\paragraph{Content Categorization \& Recommendation:} 
Our model estimates two key outputs: the evolving user interests $\{U\}^{t=1:\tau}$, and the underlying content attributes $V$. Both of them can be leveraged by digital media firms to improve their content offerings in a data-driven fashion. The content factor matrix $V$, which captures the latent content attributes, can be, for instance, used to categorize content on a news publisher's website. Typically, this categorization of news articles into a set of pre-defined categories or topics, e.g., sports, politics, business, is done manually by editors. This process could be automated in a data-driven fashion by using the $V$ matrix to classify news stories into existing categories and generate new categories, such as the content on social issues or content high in emotional valence. A firm can also adopt a hybrid approach to news categorizations and refine the editorial classifications based on our model estimates. Similarly, the temporally smoothed user factor $U^t$ can be used to generate personalized content recommendations. The most straightforward such system can be constructed by finding the news article embeddings closest in the d-dimensional space to the user embeddings $r_i^t$ and recommending such stories to the user.

However, there is a caveat that any manager implementing our suggestions needs to consider. There is no random variation in the consumption data, so it is hard to assess users' responsiveness to any recommendation or categorization performed using our model. The empirical evaluations in this paper only assess the predictiveness of the representations learned by our model. Though, in general, predictive power is correlated with the metrics determining the success of such recommendation or categorization systems.

\subsection{External Validity}
In this paper, we proposed a neural matrix factorization modeling approach to extract nonlinear patterns from text data to infer customers' evolving interests. We apply our method to model news consumption data from the Boston Globe's website. As we saw in  Figure~\ref{fig:wordcloud}, Globe's news coverage does slant towards the geographical area it serves. However, our model did not make any modeling assumptions specifically tuned to Globe or news consumption more generally. So, without making any changes, our approach can be used to model other types of textual data, for instance, various types of user-generated content, e.g., online reviews, chats, or searches.

Building on this rationale, our method can also be used to model other types of high-dimensional consumption data. By making a few changes, our approach can be used to extract nonlinear patterns from image or video data, for instance. The crucial difference in inferring dynamics user interests from visual data would be in the type of input embeddings ${E}_x$ used. Input embeddings for image or video data would need to exploit the spatial proximity\footnote{It has been shown that good visual features exhibit spatial proximity~\citep{goodfellow2016deep}.} of the input data. Once we have visual embeddings, then, the rest of the modeling can proceed as currently. One could also imagine using our model to infer consumers' dynamic interests based on their purchases of supermarket items\textemdash a common modeling context in marketing. However, it is unclear if modeling nonlinearities in such contexts will give significant improvements over existing methods such as logit models since the input data already sits in a low-dimensional space.

This paper models changes in customers' consumption behavior in online news. In many other marketing contexts such as retailing or supermarket purchases, this is often equivalent or at least assumed equal to modeling changes in demand-side consumer interests. However, in our context, a potent supply-side mechanism exists that is consistent with observed behavior. It is represented by external factors that change the supply of various kinds of news stories. In reality, online news consumption behavior is probably driven by both supply-side and demand-side factors. We do not model the supply-side in this paper and instead take the news content as exogenously determined each period. The question of disentangling supply and demand is an important one, but it is beyond the scope of this paper. The main focus and hence the contribution of this paper is in extracting nonlinear patterns from high-dimensional text data and modeling the associated habit formation in content consumption. If indeed a marketing application arises that requires modeling both supply and demand-side of news consumption, then our model can be used as a small module to extract nonlinear patterns from high-dimensional data inside a larger economic model. Modeling user behavior itself is sufficient for many predictive customer analytics applications.

\subsection{Conclusion \& Limitations}
We are living in the age of an information deluge. Firms are overwhelming customers with highly intrusive advertisements, emails \& coupons since they lack reliable estimates of customer interests. It is partly due to the companies not being able to efficiently harvest economically significant signals from the copious swathes of clickstream data and partly due to their inability to collect relevant data in the first place~\citep{melaMoorman}. Analytics approaches like ours can help firms efficiently unravel managerially relevant customer insights from high-dimensional text data. And, hence they possess the potential to move the firms towards their goal of tapping into customers' minds and increasing the relevance of their messages and content offerings.

That said, our framework is not without limitations. First, we model only one kind of customer digital footprints \textemdash news consumption. Future work should model other kinds of data, e.g., online search history, comments, reviews, and different types of UGC. Further, it is also a fruitful direction to propose new neural net models for these data to answer important marketing and customer analytics questions. To the best of our knowledge, the use of deep learning and neural net models in marketing research is still an under-explored area of study. The significant breakthroughs made in the last decade in the estimation and scalability of these families of models make a compelling case to employ them for modeling customer and firm outcomes from large datasets. Second, future work could go beyond the bag-of-words assumption we made while modeling the textual content. It could, for instance, use convolutional neural networks (CNN) or attention mechanisms to model the relative importance of different words in the consumed content. The magnitude of the economic impact of these methodological choices is an empirical question and is tough to predict. Third, we only model the demand-side and assume that consumers' consumption patterns are driven only by their consumption in previous periods. It is an exciting avenue of future research to model the interaction of content availability with readers' consumption interests. We hope our work will inspire future research to overcome these limitations in pushing the limits of our understanding of the dynamics of online content consumption.

\bibliography{targeting}
\bibliographystyle{plainnat}

\end{document}